%% file: Manuscript.tex
\documentclass[12pt]{article}
\usepackage{amsmath} 
\usepackage{cite}
\usepackage{url}
\usepackage{hyperref}
\usepackage{amsmath}
\hypersetup{colorlinks=true,allcolors=blue}
\usepackage{graphicx}
\usepackage{tabulary}
\usepackage[utf8]{inputenc}
\usepackage{fancyhdr}
\usepackage{amssymb}
\usepackage{amsthm}
\usepackage{placeins}
\usepackage{amsfonts}
\usepackage[all]{xy}
\usepackage{tikz}
\usepackage{verbatim}
\usepackage{caption}
\usepackage{subcaption}
\usepackage{multirow}
\usepackage{psfrag}
\usepackage{times}
\usepackage{float}
\usepackage{makecell}
\usepackage{lipsum}
\usepackage{tabularx}
\usepackage{xcolor}
\usepackage{soul}
\usepackage{lineno}
\usepackage{adjustbox}
\usepackage{bm}
\usepackage{booktabs}
\usepackage{rotating}
\usepackage{units}
\usepackage{bm}
\usepackage{textcomp,mathcomp}
\usepackage{amssymb} 
\usepackage{tabularx} 
\usepackage{booktabs} 
\usepackage{array} 
\usepackage[T1]{fontenc}

\usepackage[margin=1in]{geometry}
\usepackage{setspace}
\usepackage{authblk}      
\usepackage{hyperref}     
\usepackage{lmodern}      
\usepackage[T1]{fontenc}
\usepackage{microtype}
\usepackage[font=small,labelfont=bf]{caption}

\renewcommand\Affilfont{\normalsize\normalfont}
\makeatletter
\renewcommand\AB@affilsepx{\quad\protect\Affilfont} 
\makeatletter

\title{\bfseries In{-}Vivo Skin 3{-}D Surface Reconstruction and\\
Wrinkle Depth Estimation using Handheld\\
High Resolution Tactile Sensing}

\date{September 14, 2025}

\author[1]{Akhil Padmanabha}
\author[1,4]{Arpit Agarwal}
\author[1]{Catherine Li}
\author[2]{Austin Williams}
\author[3]{Dinesh K. Patel}
\author[1]{Sankalp Chopkar}
\author[3]{Achu Wilson}
\author[5]{Ahmet Ozkan}
\author[4]{Wenzhen Yuan}
\author[6]{Sonal Choudhary}
\author[7]{Arash Mostaghimi}
\author[1]{Zackory Erickson\textsuperscript{\dag}}
\author[1,3]{Carmel Majidi\textsuperscript{\dag}\thanks{Corresponding author.}}

\affil[1]{Robotics Institute, Carnegie Mellon University, Forbes Avenue, Pittsburgh, PA 15213, USA\\
\texttt{akhil.padmanabha@gmail.com}, \texttt{arpit15945@gmail.com}, \texttt{catheri4@andrew.cmu.edu},\\
\texttt{chopkarsankalp@gmail.com}, \texttt{zackory@cmu.edu}, \texttt{cmajidi@andrew.cmu.edu}}
\affil[2]{Department of Materials Science \& Engineering, Carnegie Mellon University, Forbes Avenue, Pittsburgh, PA 15213, USA\\
\texttt{austinwi@andrew.cmu.edu}}
\affil[3]{Department of Mechanical Engineering, Carnegie Mellon University, Forbes Avenue, Pittsburgh, PA 15213, USA\\
\texttt{dineshpa@andrew.cmu.edu}, \texttt{achuw@andrew.cmu.edu}}
\affil[4]{Computer Science, University of Illinois Urbana{-}Champaign, Siebel Center for Computer Science, Urbana, IL 61801, USA\\
\texttt{yuanwz@illinois.edu}}
\affil[5]{Department of Electrical and Computer Engineering, Carnegie Mellon University, Forbes Avenue, Pittsburgh, PA 15213, USA\\
\texttt{aozkan@andrew.cmu.edu}}
\affil[6]{Department of Dermatology, University of Pittsburgh Medical Center, Fifth Avenue, Pittsburgh, PA 15213, USA\\
\texttt{choudharys@upmc.edu}}
\affil[7]{Department of Dermatology, Brigham \& Women's Hospital, Longwood Avenue, Boston, MA 02115, USA\\
\texttt{amostaghimi@bwh.harvard.edu}}

\begin{document}
\maketitle

{\renewcommand{\thefootnote}{\fnsymbol{footnote}}
\footnotetext[2]{Denotes equal advising.}
\renewcommand{\thefootnote}{\arabic{footnote}}}


\begin{abstract}

Three-dimensional (3-D) skin surface reconstruction offers promise for objective and quantitative dermatological assessment, but no portable, high-resolution device exists that has been validated and used for depth reconstruction across various body locations. We present a compact 3-D skin reconstruction probe based on GelSight tactile imaging with a custom elastic gel and a learning-based reconstruction algorithm for micron-level wrinkle height estimation. Our probe, integrated into a handheld probe with force sensing for consistent contact, achieves a mean absolute error of 12.55 µm on wrinkle-like test objects. In a study with 15 participants without skin disorders, we provide the first validated wrinkle depth metrics across multiple body regions. We further demonstrate statistically significant reductions in wrinkle height at three locations following over-the-counter moisturizer application. Our work offers a validated tool for clinical and cosmetic skin analysis, with potential applications in diagnosis, treatment monitoring, and skincare efficacy evaluation.

\end{abstract}


\section{Introduction}\label{introduction}
Three-dimensional (3-D) reconstruction of the human skin surface holds broad potential in dermatology. Clinically, it may aid in diagnosing and monitoring textural changes in conditions such as eczema and psoriasis as well as in evaluating treatment efficacy. In cosmetology, it can reveal insights into skin aging, sun damage, and the effects of skincare products. Traditionally, beyond visual inspection, dermatologists rely on manual palpation~\cite{cox2006palpation, cox2007literally, punj2014palpation} and dermatoscopes~\cite{errichetti2016dermoscopy, marques2012role, choi2014skin} to assess skin texture. Portable 3-D reconstruction tools offer a promising path towards more objective and precise assessment, but no handheld probe with high spatial and depth resolution has been validated for use across body locations ~\cite{theek2020surface, dzwigalowska2013preliminary, kottner2013comparison, bloemen2011objective, langeveld2022skin}. As a result, gaps remain in the literature, particularly a lack of quantitative data on wrinkle depth across different body locations with most prior studies having focused only on the wider, deeper wrinkles found on the face~\cite{nemoto2002three, tsukahara2011relationship, cula2013assessing, carlos2023detection}.

Using photometric stereo techniques to convert tactile images into 3-D height maps~\cite{johnson2009retrographic}, GelSight provides a compact and precise solution with micron-level accuracy. GelSight sensing has already shown utility in biomedical applications such as prostate stiffness estimation~\cite{di2024using} and lump detection~\cite{jia2013lump}. Previous work has also demonstrated GelSight’s value in industrial applications, such as inspecting manufactured components~\cite{lindberg2021handheld, agarwal2023robotic}, robotic manipulation~\cite{yuan2017gelsight, padmanabha2020omnitact, lambeta2020digit, abad2020visuotactile}, material hardness estimation~\cite{yuan2017shape}, and forensics~\cite{beatty2024contrasting, lilien2015applied}. Commercial GelSight systems developed by GelSight Inc. exist but require precisely molded elastomer gels and controlled illumination to achieve high accuracy, raising costs and subsequently limiting broader adoption. Additionally, as these probes are used only with rigid objects, their gels are designed to be less elastic to increase durability, which can cause significant deformation when used on deformable surfaces like human skin. Compact and low-cost GelSight designs that use RGB lighting enable portability and ease of deployment. However, the compact form factor relies on shallow and uneven LED illumination, resulting in spatially varying lighting and indentation artifacts that can be mitigated with appropriate algorithm design.


\begin{figure}[!h]
  \centering
  \includegraphics[width = 0.8\textwidth]{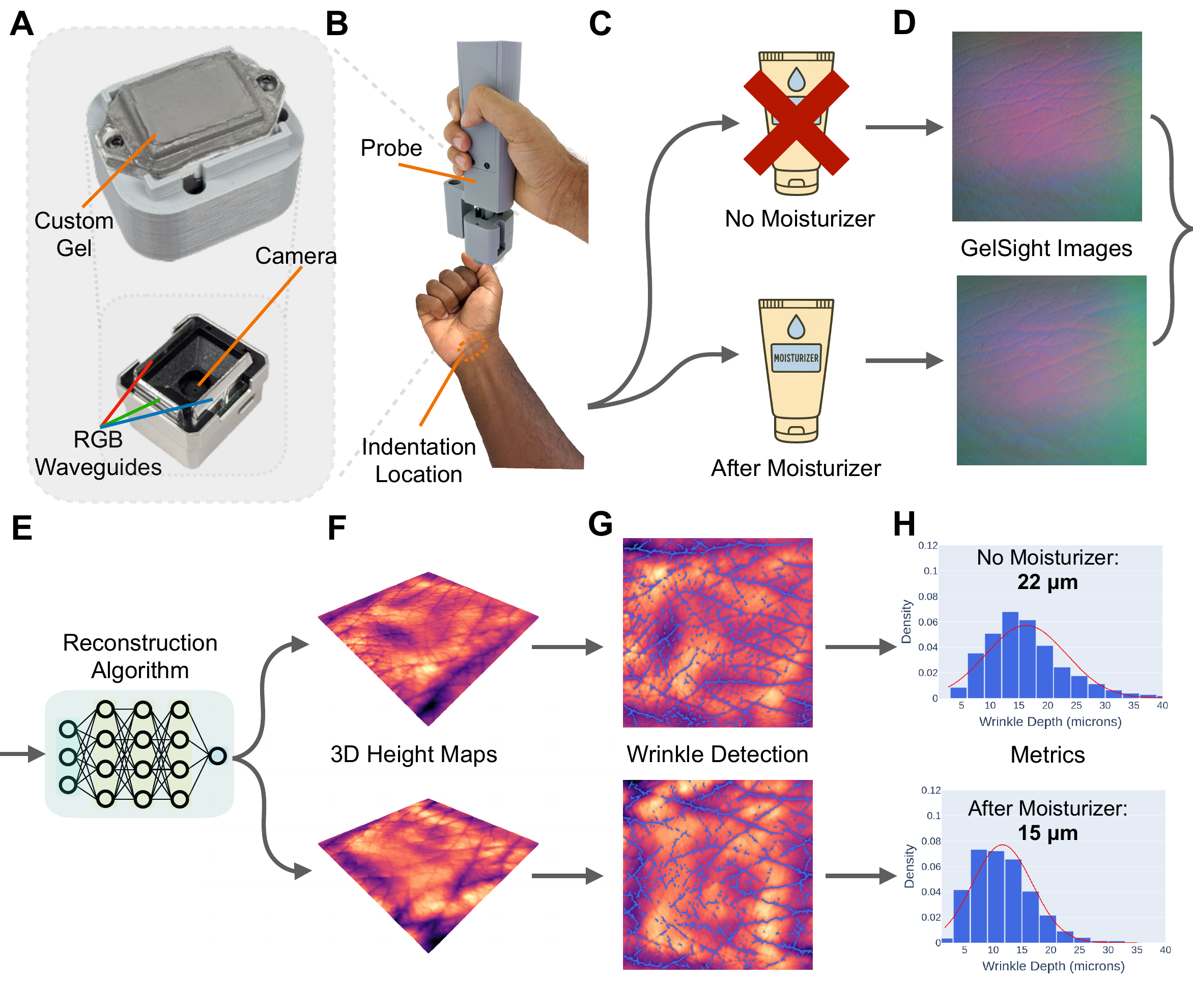}
  \caption {Skin 3-D Reconstruction Pipeline Overview. {\textbf{A.} Bottom: The GelSight Mini sensor is shown without the gel cartridge. The sensor has a high-resolution camera and three waveguides that inject red, green, and blue (RGB) light into the gel. Top: Our custom elastic gel is shown attached to the GelSight Mini sensor. \textbf{B.} Our custom 3-D printed probe is shown integrated with the GelSight sensor, custom gel, and a load cell for measuring normal force. \textbf{C.} We collect GelSight readings before and after moisturizer application. \textbf{D.} GelSight images from the wrist are shown before and after moisturizer application. \textbf{E.} Our reconstruction algorithm converts GelSight images into 3-D height maps. \textbf{F.}3-D Height Maps are shown for before and after moisturizer application for the wrist. \textbf{G.} Our wrinkle detection algorithm identifies wrinkle valleys, shown in blue overlaid on the height map. \textbf{H.} Normalized histograms of wrinkle depth are shown for before and after moisturizer application. Metrics on wrinkle depth can be automatically calculated, with the 80th percentile wrinkle depth shown. In this case, moisturizer application results in a 7 micron reduction in wrinkle height from 22 micron to 15 micron.}}
  \label{fig:overview}
\end{figure}

Motivated by the limitations of commercially available devices and existing compact GelSight technologies for skin surface height reconstruction and wrinkle depth estimation, we propose 3-D reconstruction, skin wrinkle detection, and skin wrinkle height estimation methodology and algorithms using compact GelSight sensors. We first design highly elastic gels, seen in Fig.~\ref{fig:overview}A, to reduce skin distortion during contact, enabling accurate capture of skin features. We then introduce a learning-based reconstruction algorithm that achieves a mean absolute error (MAE) of 12.55 $\mu$m on wrinkle-like test objects, allowing precise measurement of fine skin textures in any skin location on the body, exceeding the capabilities of current commercially available devices. From the resulting GelSight-generated height maps, we develop methods to detect and quantify skin wrinkles. We integrate the compact sensor and custom gel into a probe, shown in Fig.~\ref{fig:overview}B, with an embedded load cell to control the applied normal force, and evaluate its performance in measuring skin wrinkle depth.

In a human study of 15 participants without skin disorders, we use our device and methods to quantify wrinkle depth across multiple body locations, providing the first set of metrics for many of these areas in the literature. We further demonstrate the ability to detect reductions in wrinkle depth at three body locations following the application of an over-the-counter moisturizer, with statistically significant differences observed before and after moisturization for all locations. We show our full pipeline in Fig.~\ref{fig:overview}, including GelSight images from the wrist before and after moisturizer, corresponding reconstructed 3-D height maps, wrinkle detection, and estimation of wrinkle height metrics from these height maps. Unlike existing portable dermatological probes, which are typically single-purpose and lack micron-level precision, our device and methodology offer the potential for a wide range of future applications both in clinical and home settings.


\begin{figure}[t!]
  \centering
  \includegraphics[width = \textwidth]{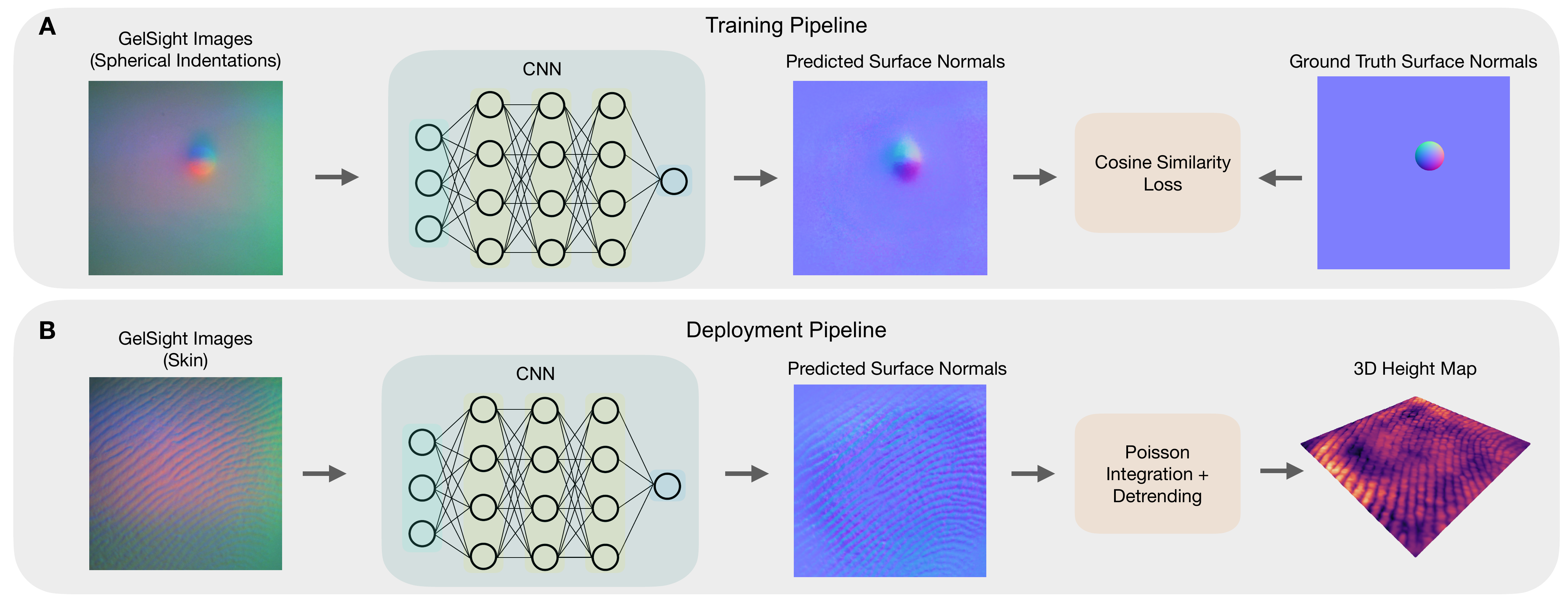}
  \caption{Training and Deployment Pipelines. \textbf{A.} Training Pipeline. GelSight images of spherical indentations with corresponding ground truth surface normals are used to train a convolutional neural network (CNN) with cosine similarity loss for prediction of surface normals. \textbf{B.} Deployment Pipeline. The trained CNN inputs GelSight skin images and outputs predicted surface normals. These surface normals are converted to 3-D height maps using poisson integration and are detrended using a 2-D high pass filter. The skin image displayed is a fingerprint.} 
  \label{fig:training_pipeline}
\end{figure}

\section{Results}\label{results}

\subsection{Gel Elasticity}
\label{sec:gel_elasticity_results}
In order to minimize deformation of the skin during contact with the GelSight sensor, we design custom elastic gels as detailed in Section~\ref{sec:gel_cartridge_design}. Our gel is attached on the GelSight Mini sensor as seen in Fig.~\ref{fig:overview}A. We use two gels produced in separate manufacturing batches, referred to as Gel 1 and Gel 2.

We use an Instron tensile testing machine, shown in Fig.~\ref{fig:methods}A, equipped with a 50 N load cell (1\% resolution) and a spherical indenter to estimate the Young's modulus of one of our two custom manufactured gel cartridges, Gel 1, using Hertzian contact theory. Additional methodology details can be found in Section~\ref{sec:gel_characterization}.

Force-displacement curves from three repeated indentation trials, performed using a 1.5 mm radius indenter indenting 2 mm into Gel 1, are shown in Fig.~\ref{fig:methods}B. The estimated Young's moduli from non-linear least squares regression for Trials 1, 2, and 3 are 124.44 kPa, 129.55 kPa, and 132.74 kPa, respectively. The average Young's modulus across trials is 128.91 kPa.

\subsection{Algorithm Performance}
\label{sec:algorithm_performance_results}

Our goal is to reconstruct a 3-D height map from GelSight RGB images. As seen in Fig.~\ref{fig:training_pipeline}B, our pipeline consists of two steps: 1. A convolutional neural network (CNN) to estimate surface normal vectors given the GelSight image input and 2. Poisson integration to estimate height from the predicted surface normals and a 2-D high-pass filter to detrend the height map. Our outputted height maps have dimensions of 1300 by 1300, corresponding to a physical area of approximately 1 cm by 1 cm.

\begin{figure}[t!]
  \centering
  \includegraphics[width = \textwidth]{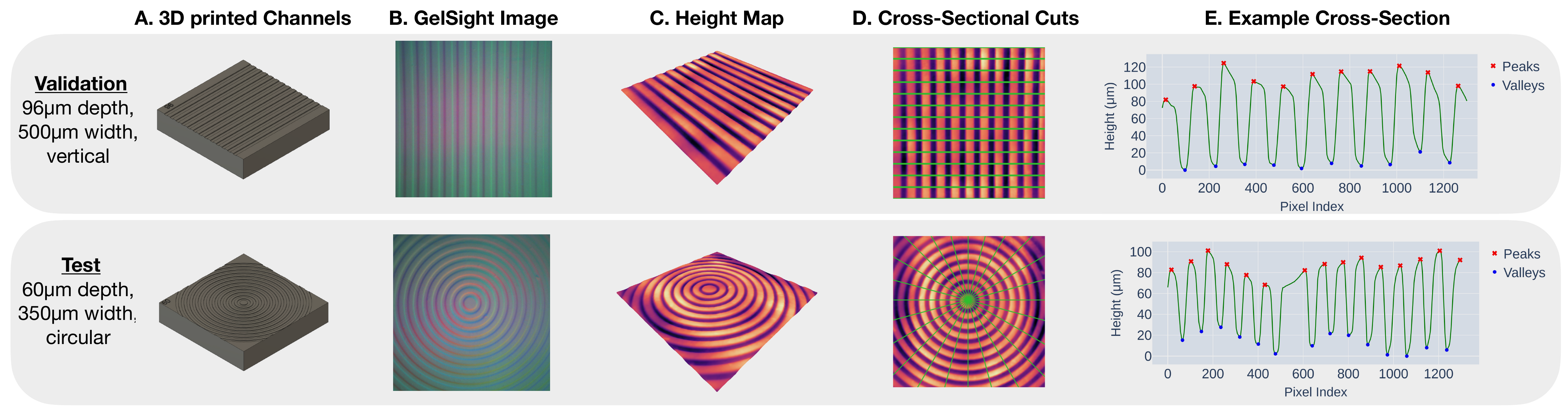}
  \caption{Evaluation Pipeline. \textbf{A.} Renderings of a representative straight channel validation object and a circular channel test object. \textbf{B.} Corresponding GelSight images. \textbf{C.} Reconstructed height maps of the two objects. \textbf{D.} The height maps are shown with green lines representing the cross-sectional cuts used to extract metrics. \textbf{E.} Two example cross-sections are shown with detected peaks and valleys used to automatically extract channel heights. }
  \label{fig:validation_pipeline}
\end{figure}

\begin{figure}[t!]
  \centering
  \includegraphics[width = \textwidth]{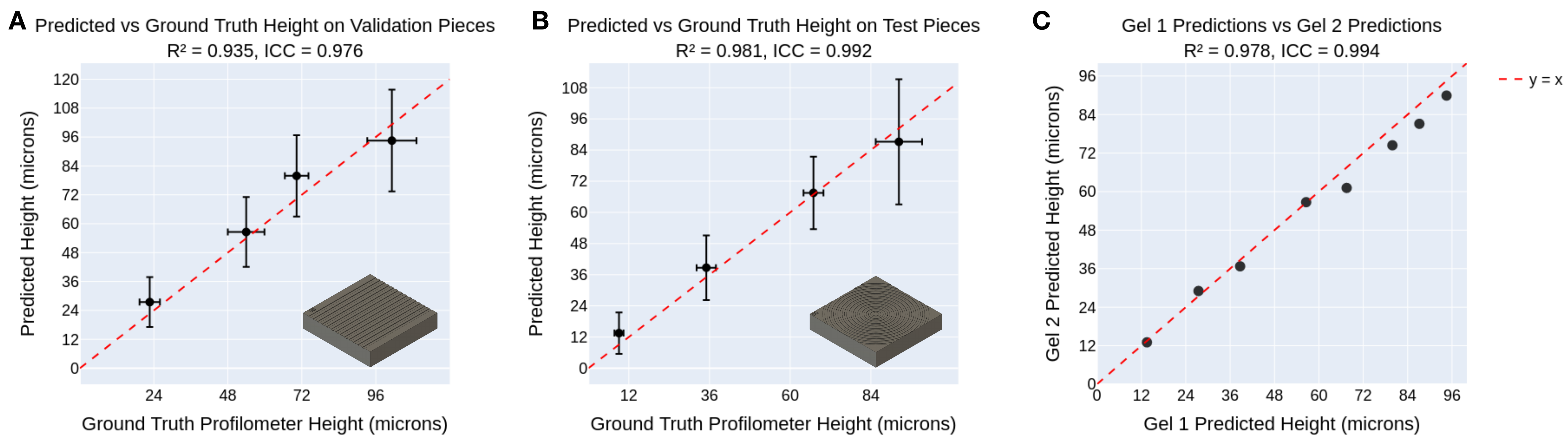}
  \caption{Sensor Accuracy and Precision on Validation and Test Objects. \textbf{A.} Mean predicted height vs. ground truth height is shown for the four straight channel validation objects. Results are averaged across Gel 1 and Gel 2. Error bars represent 1 standard deviation from the mean. \textbf{B.} Mean predicted height vs. ground truth height is shown for the four circular channel test objects. Results are averaged across Gel 1 and Gel 2. Error bars represent 1 standard deviation from the mean. \textbf{C.} Gel 1 mean predicted height vs. Gel 2 mean predicted height is plotted for all test and validation objects to assess precision of the sensors and algorithms. We include $R^2$ and ICC values for all plots, and show the $y = x$ line as a reference to indicate perfect agreement between the values on the two axes.}
  \label{fig:validation_test_results}
\end{figure}

As detailed in Section~\ref{sec:validation_test_sets}, we 3-D print 4 rigid validation and 4 rigid test pieces that consist of channels in straight-line and circular configurations, respectively. The 4 rigid validation pieces have channels of designed depth 24, 48, 72, and 96 micron with designed channel width of 500 micron while the 4 rigid test pieces have channels of designed depth 12, 36, 60, and 84 micron with designed channel width of 350 micron. To ensure that our 3-D printed objects are accurate, we use a 3-D optical profilometer to measure several cross sections of the objects. We use a peak and valley detection algorithm on the cross sections and find the difference in height between adjacent peaks and valleys to estimate the height of each channel. By doing this across all cross sections, we can get the mean and standard deviation of the channels across the object. The mean value is called the ground truth profilometer height, which we can use to compare our GelSight measurements with. The profilometer measurements for the validation objects yielded mean depths of $22.65 \pm 3.30$, $53.95 \pm 2.80$, $70.31 \pm 4.45$, and $101.22 \pm 9.19 \,\mu$m for the designed depths of 24, 48, 72, and 96 $\mu$m, respectively, while the test objects yielded mean depths of $9.09 \pm 1.35$, $35.05 \pm 2.86$, $66.92 \pm 2.95$, and $92.34 \pm 6.90 \,\mu$m for the designed depths of 12, 36, 60, and 84 $\mu$m.  

Similarly, we can automatically extract channel depth estimates from our reconstructed and detrended height maps from the GelSight sensor. In Fig.~\ref{fig:validation_pipeline}, we visually show the process of extracting these metrics from the validation and test pieces. Further information on this methodology is included in Section~\ref{sec:validation_test_sets}.

We run a series of hyperparameter sweeps using only Gel 1. We select the CNN algorithm which achieves the highest performance (lowest mean absolute error) on the validation pieces. We calculate the mean absolute error using the designed depths, since these values closely align with the ground-truth profilometer measurements, included in the Supporting Information Table S7 and S8. We use the test pieces to show generalization of our method to different channel shapes, widths, and depths. On the validation pieces, averaged across both gels and all 4 objects and using the designed ground truth values, our algorithm achieves a validation MAE of $14.01 \pm 11.41$ $\mu$m. On the test pieces, the average MAE across both gels is $12.55 \pm 11.35$ $\mu$m. Specifically, Gel 1 achieves a validation MAE of $13.83 \pm 11.15$ $\mu$m and a test MAE of $14.67 \pm 13.49$ $\mu$m, while Gel 2 achieves a validation MAE of $14.19 \pm 11.67$ $\mu$m and a test MAE of $10.42 \pm 9.21$ $\mu$m. 

We summarize algorithm performance on our validation and test sets in Fig.~\ref{fig:validation_test_results}A and B. For these plots, we average the results across both Gel 1 and Gel 2, with all data points provided in Table S7 and S8 in the Supporting Information. The GelSight reconstructions for the validation objects yielded mean depths of $27.45 \pm 10.36$, $56.56 \pm 14.51$, $79.88 \pm 16.92$, and $94.51 \pm 21.13 \,\mu$m for the designed depths of 24, 48, 72, and 96 $\mu$m, respectively, while the test objects yielded mean depths of $13.50 \pm 7.96$, $38.70 \pm 12.46$, $67.51 \pm 13.96$, and $87.17 \pm 24.10 \,\mu$m for the designed depths of 12, 36, 60, and 84 $\mu$m. Both Fig.~\ref{fig:validation_test_results}A and B show strong correlation between ground truth profilometer height and predicted height for all objects. For the validation pieces, our algorithms achieve a coefficient of determination ($R^2$) of 0.935 and an Intraclass Correlation Coefficient (ICC) of 0.976, while for the test pieces, we get an $R^2$ of 0.981 and an ICC of 0.992. 

In Fig.~\ref{fig:validation_test_results}C, to analyze repeatability of our sensor with different gels, we plot the mean predicted height for each of the 4 validation objects and the 4 test objects for Gel 1 versus Gel 2. We obtain an $R^2$ of 0.974 and ICC of 0.993, showing strong repeatability of our sensor and algorithm, even with two gels. 

\subsection{Characterization of Skin Wrinkles}
\label{sec:skin_wrinkle_results}

\begin{figure}[p] 
  \centering
\includegraphics[width=0.7\textwidth]{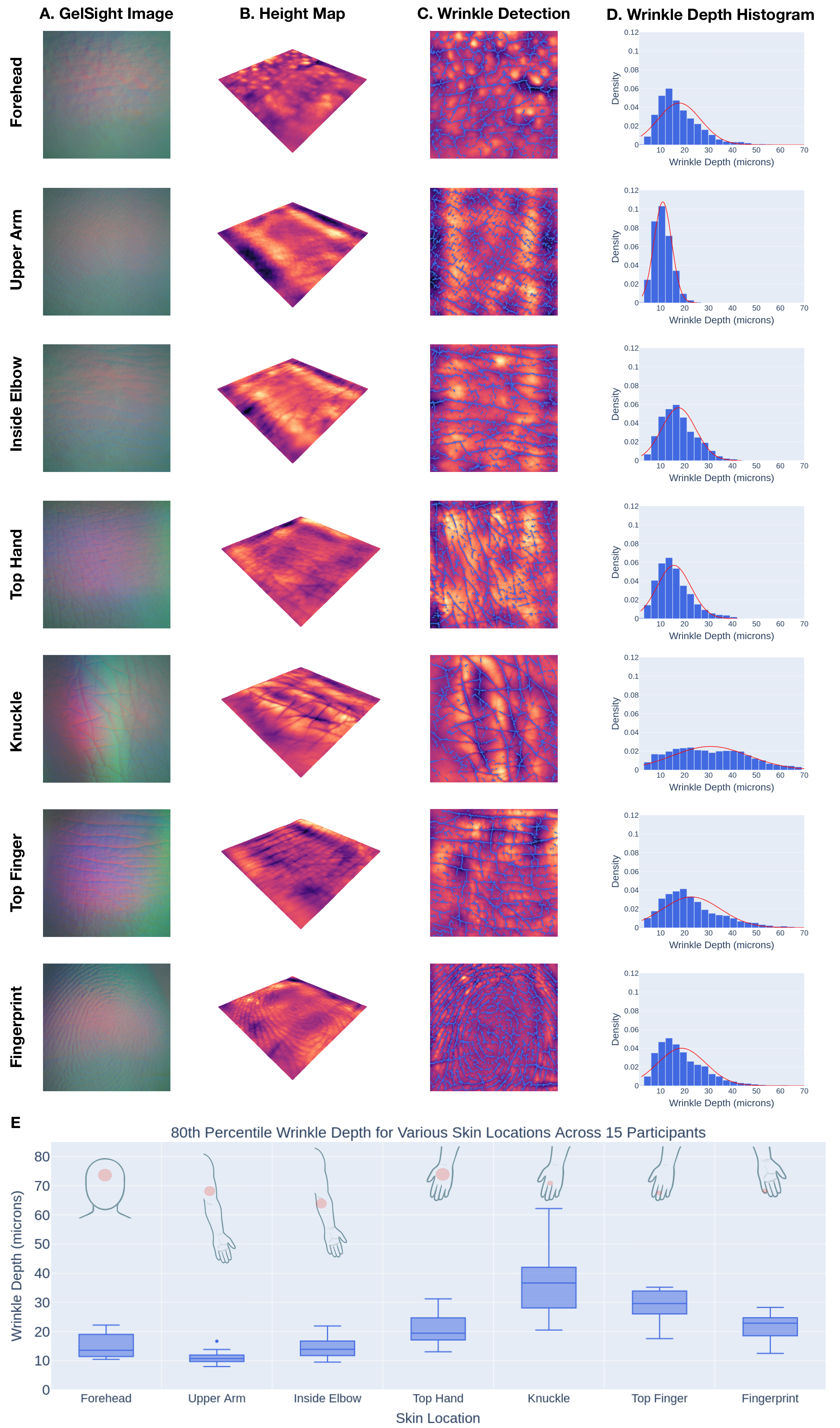} 
\caption{3-D reconstruction, Wrinkle Detection, and Wrinkle Depth Estimation in Various Locations on the Human Body. \textbf{A.} GelSight images from seven skin locations: forehead, upper arm, inside elbow, top of hand, knuckle, top of finger, fingerprint. \textbf{B.} Corresponding 3-D height maps are shown. \textbf{C.} The detected wrinkle valleys (troughs) are shown in blue overlaid over the height map. 
\textbf{D.} Normalized histograms of wrinkle depth are shown for each location, with Gaussian fits overlaid in red. \textbf{E.} Boxplots of the 80th percentile wrinkle depths across all 15 participants are shown with clear differences seen across the various locations. 
}
\label{fig:skin_various_locations}
\end{figure}
As detailed in Section~\ref{sec:wrinkle_estimation_methods}, we introduce methods for detection of skin wrinkles from 3-D height maps and estimation of wrinkle height. These methods input in the 3-D height maps and output valleys that correspond to where wrinkles are located. Using these detected valleys, we can estimate wrinkle depths throughout the height map using surrounding peaks. 

To further validate our probe, reconstructions algorithms, and wrinkle detection methods, we conducted a human study with participants with no skin disorders. We detail our human study procedure in Section~\ref{sec:human_study_procedure}. Fifteen participants (9 male, 6 female; sex and gender by self-report) were recruited for the study, with a mean age of 23.20 years (SD = 3.91, range = 18–31). Fourteen participants self-identified as Asian and one as African-American. No participants reported having excessive wrinkles at any of the measured body locations. Fitzpatrick skin types were distributed as follows: Type III (2 participants), Type IV (9 participants), Type V (2 participants), and Type VI (2 participants). 12 out of 15 participants did not apply any creams, lotions, or moisturizers to their skin within four hours prior to the study. One participant reported applying a product to their elbow and palm, another to their elbow and wrist, and a third to their palm. Gel 1 was used for 6 participants while Gel 2 was used for the remaining 9 participants. 

Wrinkle depth can vary depending on whether the skin is relaxed or stretched, particularly around joints such as the elbow. To ensure consistency across participants, we standardize joint positioning during all measurements as described in Section~\ref{sec:human_study_procedure}. The specific body locations are illustrated in Fig.\ref{fig:skin_various_locations}E and Fig.\ref{fig:before_after_cream}E. We present all raw data from our human study in Table S6 in the Supporting Information. 

In Fig.~\ref{fig:skin_various_locations}A and B, we present a subset of GelSight images collected in our human study and corresponding reconstructed 3-D height maps from 7 skin locations: forehead, upper arm, inside elbow, top of hand, knuckle, top of finger, and fingerprint. In Fig.~\ref{fig:skin_various_locations}C and D, we visually illustrate the results of our wrinkle detection algorithm along with histograms of the estimated wrinkle depths across the reconstructed surfaces. In Fig.~\ref{fig:skin_various_locations}E, we show boxplots of 80th percentile wrinkle depth for each location for all participants, providing a representative measure of overall wrinkle depth while reducing the influence of numerous micro-wrinkles.

Across all participants, the knuckle region exhibited the highest mean 80th percentile wrinkle depth at 35.90 micron, followed by the top of the finger (28.46 micron) and the fingerprint region (21.54 micron). The top of the hand also showed relatively high values with a mean of 21.12 micron. In contrast, lower mean wrinkle depths were observed in the upper arm (11.09 micron), inside elbow (14.35 micron), and forehead (14.88 micron). Full summary statistics including median, minimum, and maximum values are provided in the Supporting Information Table S4.

\subsection{Effect of Moisturizer on Wrinkle Depth}
\begin{figure}[p] 
  \centering
\includegraphics[width=0.7\textwidth]{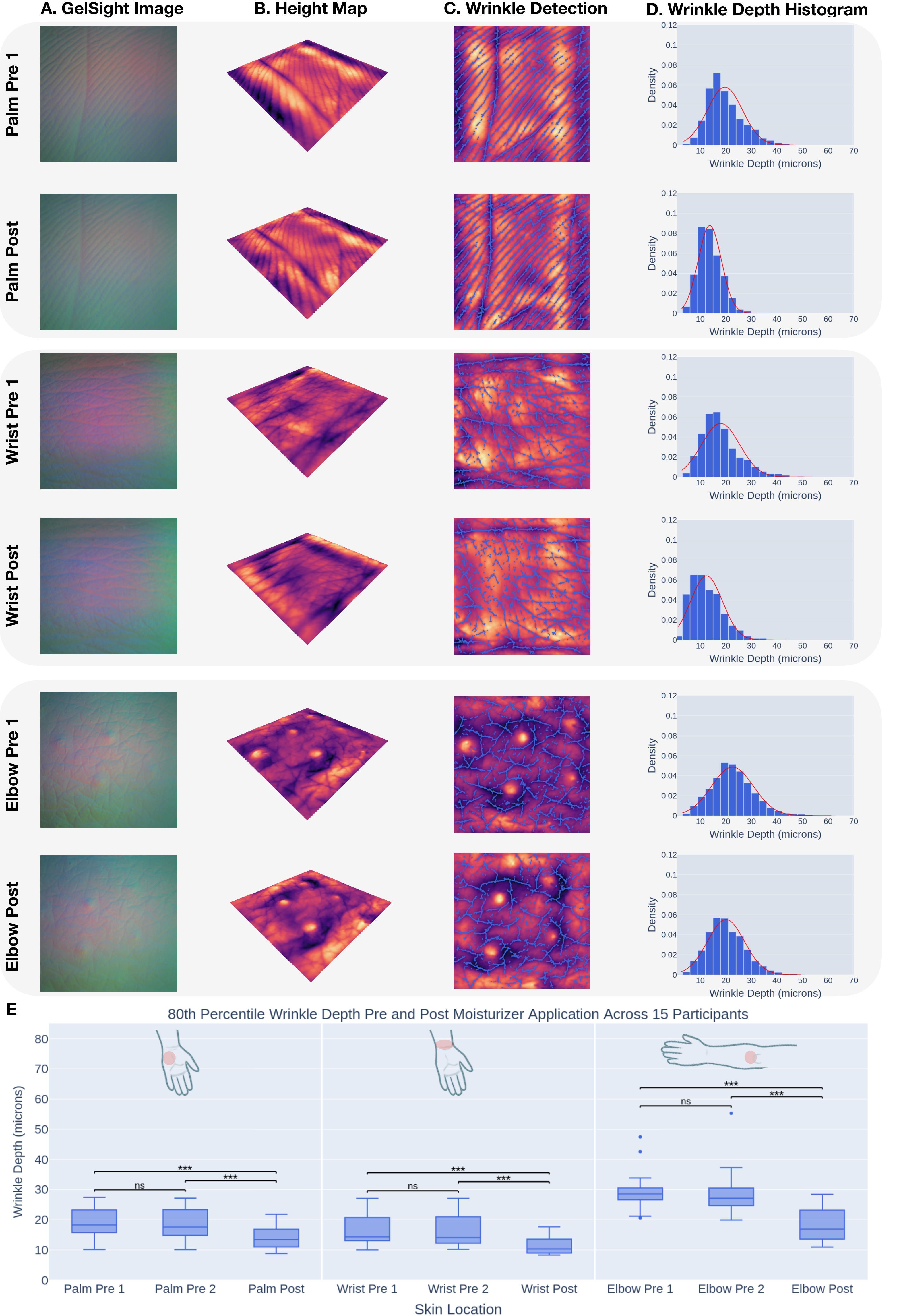} 
  \caption{Effect of Moisturizer on Wrinkle Depth. \textbf{A.} GelSight images from three skin locations: palm, wrist, and elbow. Two readings (Pre 1 and Pre 2) are taken pre-moisturizer application and 1 reading (Post) is taken after application. \textbf{B.} Corresponding 3-D height maps are shown. \textbf{C.} The detected wrinkle valleys (troughs) are shown in blue overlaid over the height map. 
\textbf{D.} Normalized histograms of wrinkle depth are shown for each location, with Gaussian fits overlaid in red. The post-application Gaussian fits are shifted left relative to the pre-application curves, indicating reduced wrinkle depth. \textbf{E.} Boxplots of the 80th percentile wrinkle depths across all 15 participants are shown for the two pre moisturizer readings and one post moisturizer reading. Asterisks denote statistical significance: * $p < 0.05$, ** $p < 0.01$, *** $p < 0.001$. No statistical difference was found between the two pre-application readings, indicating sensor repeatability. Both pre-application readings showed strong statistical differences from the post-application reading (Wilcoxon signed-rank test with Bonferroni correction for 3 comparisons), demonstrating the sensor's ability to detect changes in wrinkle depth due to moisturizer across participants.  
}
\label{fig:before_after_cream}
\end{figure}

In Fig.\ref{fig:before_after_cream}A, we present a subset of GelSight images and corresponding 3-D height maps captured before and after moisturizer application in our human study. We also include the outputs of our wrinkle detection algorithm and histograms of estimated wrinkle depths across the reconstructed surfaces. Fig.\ref{fig:before_after_cream}B displays boxplots of the 80th percentile wrinkle depth for each location (palm, wrist, and elbow) across all participants, including two pre-moisturizer readings, named Pre 1 and Pre 2, and one post-moisturizer reading, named Post.

Across all three locations, we observe consistent reductions in 80th percentile wrinkle depth following moisturizer application. On average, the palm shows a decrease from 19.10 micron (Pre 1) and 18.77 micron (Pre 2) to 13.90 micron (Post). The wrist exhibits a reduction from 16.73 micron and 16.34 micron to 11.33 micron, and the elbow from 29.88 micron and 29.05 micron to 18.57 micron.

To evaluate statistical significance, we apply the non-parametric Friedman test to compare the three time points at each location. The test reveals statistically significant differences for the palm ($\chi^2 = 19.73$, $p = .0001$), wrist ($\chi^2 = 22.53$, $p < .0001$), and elbow ($\chi^2 = 21.73$, $p < .0001$). For locations with significant overall effects, we perform post-hoc Wilcoxon signed-rank tests with a Bonferroni correction factor of 3. All statistical testing values are located in the Supporting Information Table S3. These tests indicate no significant difference between the two pre-moisturizer readings and statistically significant reductions in wrinkle depth between both pre readings and the post reading at all locations. These results demonstrate both the repeatability of our pipeline and the measurable effect of moisturizer application on reducing wrinkle depth.

\section{Discussion}
We present a compact, high-resolution 3-D skin reconstruction probe and learning-based algorithm capable of estimating micron-level wrinkle depth using GelSight tactile sensing. Our system is accurate, achieving just 12.55 micron error, and repeatable across different gels. It additionally generalizes to a wide range of skin topographies at different locations of the body and reliably detects changes in surface morphology following topical treatments. This is the first handheld in-vivo 3-D skin reconstruction method validated on test objects and through a human study. Existing skin measurement methods fall into direct and indirect categories~\cite{fischer1999direct}. Indirect methods, such as silicone replica-based profilometry using optical, confocal, or laser systems~\cite{cook1982quantification, courage2025visioline}, offer high accuracy but require bulky lab setups and are unsuitable for in-home use. In contrast, direct methods—--typically integrated into handheld probes---lack accurate depth (z-axis) estimation, limiting their utility for applications requiring precise surface assessment~\cite{langeveld2022skin}. The most common direct approach, surface evaluation of living skin (SELS), uses integrated RGB cameras to assess roughness and wrinkling~\cite{fischer1999direct, courage2024visioscan}. However, SELS is not validated for 3-D reconstruction and has limited depth resolution (e.g., 50 µm for the Visioscan VC 20plus~\cite{courage2024visioscan}), which is insufficient for detecting fine wrinkles. Optical stereo and fringe projection methods, like the PRIMOS system\cite{roques2003primos}, offer 3-D measurements but require mounts to limit motion in subjects, restricting their use mostly to facial wrinkles~\cite{bloemen2011objective, trojahn2015reliability}. Additionally, SELS and PRIMOS rely on optical imaging of the skin and have not been rigorously validated for diverse skin tones, raising concerns about generalizability. In contrast, while GelSight is an optical method, it uses opaque gel coatings and thus is completely unaffected by skin pigmentation. A direct comparison of our method and its performance with three commercially available systems is provided in Table S2 in the Supporting Information.

Our custom-designed gels, characterized using indentation testing and Hertzian contact theory, achieved Young's modulus values around 129 kPa, helping minimize deformation during measurement. For comparison, skin elasticity reported in the literature varies widely from tens of kPa to over 3 MPa depending on factors such as skin location, hydration, measurement method, and the specific skin layer being assessed~\cite{van2013contact, pailler2008vivo}. While our gel ensures minimal deformation for less elastic skin, it may be less effective on highly elastic regions. To address this limitation, future work can explore the design of more elastic gels to further reduce skin deformation during contact. In parallel, efforts can focus on directly accounting for skin deformation in wrinkle depth estimation. For instance, placing a spherical object between the gel and the skin and applying Hertzian contact theory could allow for estimation of local skin elasticity. Alternatively, the skin location could be identified either manually or via a trained classifier using GelSight images or reconstructed height maps and combined with participant-specific data (e.g. age) in a lookup table to estimate elasticity and adjust measurements accordingly.

The core of our reconstruction pipeline combines a CNN-based surface normal estimation model with Poisson integration and high-pass filtering to generate detrended height maps of the skin surface. Our methods were rigorously evaluated using both straight and circular channel objects, shown in Fig.~\ref{fig:validation_pipeline}, 3-D printed at micron-scale depths, mimicking the geometry of natural skin wrinkles. On these test structures, our method achieves 12.55 micron mean absolute error (MAE) with high $R^2$ and ICC values, indicating both accuracy and repeatability. Importantly, our model generalizes well across different geometries and different gels. 

We show the utility of our methods to skin surfaces by contributing wrinkle detection algorithms. We qualitatively displayed the performance of these algorithms for detecting wrinkles in 10 locations on the human body: forehead, upper arm, inside elbow, top of hand, knuckle, top of finger, fingerprint, palm, wrist, and outside elbow. We also presented methods on using this pipeline to automatically extract histograms of wrinkle depth and other wrinkle metrics. In our human study across 15 participants and 10 body locations, we observe distinct variations in wrinkle depth based on the location on the human body. As anticipated, regions exposed to frequent movement and mechanical stress, such as the knuckles and fingers, exhibit higher wrinkle depths, while smoother areas like the upper arm show lower values. We also observe that joint orientation affects measured wrinkle depths; for instance, the outside elbow yields lower than expected values due to the skin being stretched when the elbow is bent.

Our work is the first to enable quantitative measurement of wrinkle depth in physical units (microns) across a wide range of body locations and participants. Prior studies have primarily focused on prominent facial wrinkles, such as frontal furrows (horizontal lines), glabellar lines (frown lines), and lateral canthal lines (crow's feet)\cite{nemoto2002three, tsukahara2011relationship, cula2013assessing, carlos2023detection}. While some research has examined fingerprint wrinkle depth, most do not report results in micron-scale units~\cite{huang20143d, wang2009noncontact}. One study reports wrinkle depth around 30 $\mu$m~\cite{ha2015multi}, but only for a single participant. In contrast, our system generalizes to any location where the handheld probe can be applied and is capable of detecting and estimating the depth of all wrinkles, including fine micro-wrinkles. 

Lastly, we demonstrate the ability of our probe to detect changes in skin wrinkles due to topical treatments. Across palm, wrist, and elbow sites, wrinkle depth significantly decreases following moisturizer application, with statistically significant differences confirmed using non-parametric testing. The minimal variation between the two pre-moisturizer measurements on the extracted 80th percentile wrinkle depth further underscores the repeatability of our probe and algorithms. These findings validate the potential of our system to detect subtle changes in skin hydration and elasticity and may have applications in dermatological research, cosmetics testing, and longitudinal skin health monitoring.

Overall, our results show that combining tactile imaging with learning-based surface reconstruction and wrinkle detection and estimation pipelines offers a strong approach to quantifying fine-grained skin surface properties. Future work should examine performance of our probe and algorithms in older populations or those with clinically relevant skin conditions (e.g., eczema, psoriasis). 

\section{Experimental Section}\label{methods}



\begin{figure}[t!]
  \centering
  \includegraphics[width = \textwidth]{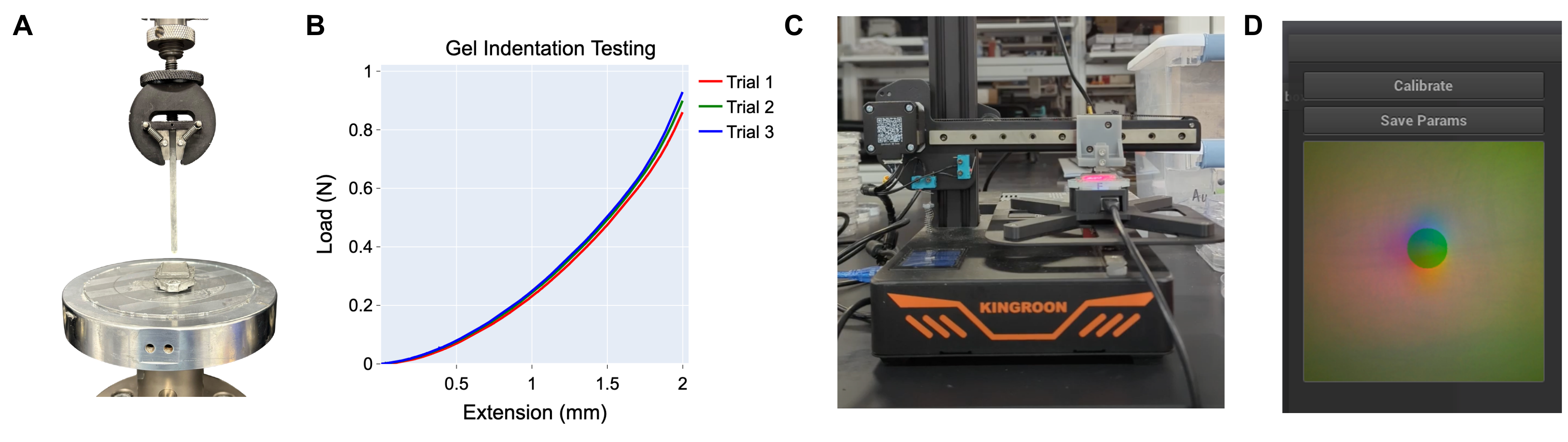}
  \caption{Gel Characterization and Calibration. \textbf{A.} Instron indentation testing setup used to estimate the elasticity of the custom gels. \textbf{B.} Load (N) versus extension (mm) curves from three indentation test trials. \textbf{C.} Our calibration indentation rig is shown with the indenter attached and the GelSight sensor, featuring the custom gel, mounted on the stage. \textbf{D.} Our indentation labeling graphical user interface (GUI) is shown which allows us to label GelSight images with corresponding ground truth surface normals. }
  \label{fig:methods}
\end{figure}

\subsection{Sensor}
We utilized the GelSight Mini Tactile Sensor from GelSight Inc. The GelSight Mini consists of a high-resolution 8 megapixel camera providing images of 2448 by 3264 pixels. Red, green, and blue light from LEDS from three of the four sides of the rectangular sensor are directed into an acrylic piece attached to the gel cartridge providing uniform illumination of the gel surface. The GelSight Mini sensor without the gel cartridge is shown in Fig.~\ref{fig:overview}A. Additional information on the sensor can be found on the GelSight Inc. website~\footnote{GelSight Inc. \url{https://www.gelsight.com/}}. 

\subsection{Gel Design}
\label{sec:gel_cartridge_design}
To minimize deformation of the soft material as the GelSight is pressed against its surface, we designed and manufactured custom gels for the GelSight Mini sensor. Our custom gel attached to the sensor is shown in Fig.~\ref{fig:overview}A. The cartridge hardware consists of a 3-D printed mount made of PLA (polylactic acid) and a clear acrylic piece (0.7 inches by 0.95 inches by 0.11 inches). First, the acrylic is press fit and glued into the 3-D-printed mount. A mold is constructed using another 3-D printed piece. The gel consists of three layers: base elastomer, powder coating, and the skin-safe elastomer. The base elastomer is made with Silicones Inc. XP-565 at a 1:22 ratio of the activator to the base. The two parts of the elastomer are mixed, degassed for 12 minutes, and poured into the mold. The elastomer is cured for 24 hrs in an oven at $55^{\circ}C$ and removed from the mold. Next, to add the opaque diffuse coating for the gel, we used solid aluminum spherical powder with particles of diameter 1 {\textmu}m from US Research Nanomaterials, Inc. The aluminum powder was first sifted using a sieve to remove powder clumps and then applied using a makeup brush to the surface and sides of the gel. Next, we mix a certified skin-safe elastomer, Smooth-On SORTA-Clear™ 18, with a solvent, Smooth-On NOVOCS™ Matte, in a ratio of 1:16.5:33 for the activator to the base to the solvent. We pour the mixture over the coated gel and leave it tilted for 5 min for the excess elastomer to flow off. Lastly, the gel is cured for 24 hrs in an oven at $55^{\circ}C$.

\subsection{Gel Mechanical Characterization}
\label{sec:gel_characterization}
We characterized the mechanical properties of our soft, skin-safe gels, described in Section~\ref{sec:gel_cartridge_design}, using a spherical indenter with radius, $R = 1.5$ mm, and an Instron tensile testing machine. The spherical indenter is 3-D printed using a Formlabs Form 3 SLA printer with Gray Pro Resin. Our setup is shown in Fig.~\ref{methods}A. Using Hertzian contact theory, we can estimate the Young's modulus of the gel, $E_{gel}$. 

For a sphere indented into a half-space, using Hertzian contact theory, we can use the following equations:
\begin{equation}
\label{eq:force_equation_unsimplified}
F =  \frac{4}{3}E^*R^{\frac{1}{2}}d^{\frac{3}{2}}
\end{equation}

\begin{equation}
\label{eq:modulus_unsimplified}
\frac{1}{E^*} = \frac{1-v_1^2}{E_1} + \frac{1-v_2^2}{E_2} 
\end{equation}

where F is the applied force, R is the radius of the spherical indenter, d is the displacement, $E_1$ and $E_2$ are the Young's moduli of the indenter and gel respectively, and $v_1$ and $v_2$ are the Poisson ratio's of the indenter and gel respectively. To simplify Eq.~\ref{eq:modulus_unsimplified}, we can assume $E_1$ is infinity as the indenter is orders of magnitude more stiff than the gel. Thus, 

\begin{equation}
\label{eq:modulus_simplified}
\frac{1}{E^*} = \frac{1-v_2^2}{E_2} 
\end{equation}

We assume $v_2 = 0.49$ as silicone elastomers can be approximated as nearly incompressible materials. Substituting Eq.~\ref{eq:modulus_simplified} into Eq.~\ref{eq:force_equation_unsimplified}, we get the following:

\begin{equation}
\label{eq:force_equation_simplified}
F =  \frac{4}{3}(\frac{E_2}{1-v_2^2})R^{\frac{1}{2}}d^{\frac{3}{2}}
\end{equation}

Lastly, using the raw data from the force-displacement curves, shown in Fig.~\ref{fig:methods}B, from the Instron machine, we can solve for $E_2$ in Eq.~\ref{eq:force_equation_simplified}, the Young's modulus of the gel, by using non-linear least squares regression. We conduct 3 total indentation tests on Gel 1, indenting 2 mm into the gel. Results from these tests are presented in Section~\ref{sec:gel_elasticity_results}.

\subsection{Calibration Indenting Rig}
\label{sec:indentation_rig}
In order to collect calibration data for each cartridge, we used a modified Kingroon KP3 FDM 3-D printer as a 3-axis indenting rig. As seen in Fig.~\ref{fig:methods}C, the sensor is mounted to the center of the 3-D printer bed using a 3-D printed mount. The printer nozzle is removed and an indenter is attached in place of the nozzle using a 3-D printed mount. We 3-D printed 3 spherical indenters with radii of 0.5, 0.875, and 1.25 mm using the Formlabs Form 3 stereolithography (SLA) printer out of Clear Resin with a layer thickness of 25 micron. 

We utilized a Python script to concurrently send commands to the calibration indenting rig and to collect data from the mounted GelSight sensor. For each indenter, 49 total indentations are collected in a grid pattern. Specifically, from the center (0, 0) x,y location on the gel, we indent at various x and y coordinates. We define two sets, $X = {(-4.5, -3, -1.5, 0, 1.5, 3, 4.5)}$ and $Y = {(-4.5, -3, -1.5, 0, 1.5, 3, 4.5)}$ and use the 49 unique combinations of these X and Y values. 

Because the gel surface may not be perfectly flat, we use a simple computer vision algorithm to align the indenter with the surface of the gel before indenting by the radius of the spherical indenter. The algorithm automatically detects the moment when the indenter is perfectly aligned with the surface of the gel by moving down at larger increments and then detecting when the indenter untouches the gel by moving up at smaller increments. This can be defined mathematically as follows:

For each unique position, (x,y) with $x \in X$ and $y \in Y$, the printer first moves in the positive z-axis to a clearance height and then to the (x,y) position. We define $I_{untouched} \in R^{300 \times 300 \times 3}$, which is the untouched median image of 5 total images taken concurrently with a crop of 300 by 300 around the indentation location. We also define $e_{untouched}$, the median of the pixel intensity mean squared error (MSE) between all unique pairwise combinations of the 5 images with a crop of 300 by 300 around the indentation location. Lastly, we define $t_t = 1.1*e_{untouched}$, the touch threshold. We use 5 images with medians to account for the measurement noise in the images. Next, we move the indenter down by 50 micron incrementally. Each time the indenter is moved, we collect 5 new images, crop them around the indentation location, and find $E_{curr} \in R^5$ which is the MSE between each image and $I_{untouched}$. Next, $e_{curr}$ is calculated using the median of $E_{curr}$ and once $e_{curr} > t_t$, we know that the indenter is touching the gel. At this point, we set $e_{final} = e_{curr}$ and we define $t_u$, the untouched threshold which is defined as $t_u = 0.7*e_{final}$. We follow a similar process to detect when the indenter untouches the gel, but using positive z-axis movements of 15 micron. When $e_{curr} < t_u$, we know the indenter is no longer touching. Finally, we indent the gel by the radius of the indenter. Using this process, we can ensure that our indentations have a depth error of at most 15 micron. The thresholds were selected by the research team through testing and visual inspection of the GelSight images. 

Using a set of digital calipers lightly pressed into the gel, we estimate the millimeter to pixel conversion factor as 0.0077 mm/pixel. Next, for each indenter, we visually examine the GelSight images from the indentation at (x,y) = (0,0). As seen in Fig.~\ref{methods}D, we visually label the center of each indentation and we use these values to automatically label the centers of the other 48 indentations. Using these center labels and the radius of the indenters converted to pixel space using the mm to pixel conversion factor, we generate ground truth surface normal images to pair with each GelSight image. An example ground truth surface normal image is shown in Fig.~\ref{fig:training_pipeline}.

\subsection{Convolutional Neural Network (CNN)}
\label{sec:CNN_details}
Our training pipeline is illustrated in Fig.~\ref{fig:training_pipeline}. Each GelSight image is cropped from its original size of 2448 by 3264 pixels to a 1500 by 1500 pixel region. We center the crop at (1224, 1482), which corresponds to a 150-pixel shift to the left along the width dimension. This cropping removes border regions with uneven RGB illumination.

Using the 144 images, 48 from each size of indenter, and the corresponding ground truth normals, we train a convolutional neural network (CNN) to estimate surface normals from the RGB GelSight images. We use a train-test split of 90-10 to partition our data into training and testing sets. The CNN input, $f \in {R^{1500 \times 1500 \times 8}}$, consists of the GelSight image, $I \in {R^{1500 \times 1500 \times 3}}$, the unindented (untouched) GelSight image, $B \in {R^{1500 \times 1500 \times 3}}$, and two positional encodings for the image, $u \in {R^{1500 \times 1500 \times 1}}$, and $ v \in {R^{1500 \times 1500 \times 1}}$. The untouched GelSight image helps the CNN identify the changes in lighting during indentation while the positional encodings help the CNN learn the variation in lighting across the entire surface of the gel. We pass our input through two convolutional blocks, each consisting of a 2D convolution layer with 64 output channels, a kernel size of 3, stride of 1, and “same” padding to preserve spatial dimensions. Each convolutional layer is followed by a batch normalization layer to stabilize training and for better generalization, and a ReLU activation function to introduce non-linearity. A final convolutional layer maps the output, $O \in {R^{1500 \times 1500 \times 3}}$, to 3 channels using the same kernel size, stride, and padding. We use cosine similarity loss defined as
\[
\mathcal{L}_{\text{cos}} = \frac{1}{\sum_{i=1}^{H} \sum_{j=1}^{W} M^{(i,j)}} \sum_{i=1}^{H} \sum_{j=1}^{W} M^{(i,j)} \left(1 - \frac{\langle \mathbf{n}_{\text{pred}}^{(i,j)}, \mathbf{n}_{\text{gt}}^{(i,j)} \rangle}{\|\mathbf{n}_{\text{pred}}^{(i,j)}\| \cdot \|\mathbf{n}_{\text{gt}}^{(i,j)}\|} \right)
\]
where \( M \in \{0, 1\}^{H \times W} \) is a binary mask indicating the loss regions, \( H \) and \( W \) are the height and width of the image (e.g., \(1500 \times 1500\)), and \(\mathbf{n}_{\text{pred}}^{(i,j)}\) and \(\mathbf{n}_{\text{gt}}^{(i,j)}\) denote the predicted and ground truth normal vectors at pixel \((i, j)\). 

The mask \( M \) is constructed as a binary matrix with 1s in a square region surrounding the indentation and an additional random subset of pixels outside this region, selected to cover a fixed percentage \( \gamma = 5\% \) of the total image area. This supervision allows the model to learn the lighting changes associated with deformations while also encouraging accurate estimation in unindented regions, which should be predicted as flat.

We implement our CNN architecture in PyTorch and train it using one NVIDIA 4090 and one 4070 GPU in parallel. Due to the high dimensionality of the input data, we use a batch size of 1 with 32 gradient accumulation steps. Training is performed using the ADAM optimizer with a weight decay of $1 \times 10^{-5}$ and an initial learning rate of $1 \times 10^{-3}$, which decays exponentially with a gamma of $0.95$. We train the model for a total of 30 epochs. We perform several hyperparameter sweeps over parameters such as kernel size (1, 3, 5, 11), number of convolutional blocks (1, 2, 3), and number of output channels per block (16, 32, 64, 128). The final model configuration corresponds to the best-performing setup on the validation objects using Gel 1. We then train a separate model for Gel 2 using the same parameters.

\subsection{Poisson Integration and Detrending}
\label{sec:poisson_integration}
As seen in Fig.~\ref{fig:training_pipeline}, during deployment, we use Poisson integration to convert surface normals into a 3-D height map. Prior works commonly use the discrete sine transform (DST)-based solvers with Dirichlet boundary conditions, which assumes the height is zero at the image boundaries~\cite{yuan2017gelsight, agarwal2025vision}. However, this assumption does not hold for our application, as the gel surface often deforms at the edges while pressing the gel against skin. Instead, we use an FFT-based Poisson solver with periodic boundary conditions. While periodicity is not physically accurate, it avoids imposing artificial constraints at the borders and we find it works well in practice since our primary interest lies in local variations in surface height rather than absolute global structure. 

After Poisson integration, we additionally detrend the height map to highlight higher frequency features by applying a 2-D high-pass filter with a cutoff frequency of 0.002, effectively removing low-frequency components such as large-scale surface curvature or tilt that is larger than 500 pixels, approximately 3.85 mm. This allows us to focus on finer geometric details and local variations in the surface, which are more informative for wrinkle detection. We crop the boundaries of our reconstructed height maps by 100 pixels on each side to remove the effects of boundary artifacts introduced by the Poisson integration solver. Our final height maps have dimensions of 1300 by 1300, corresponding to a physical area of approximately 1 cm by 1 cm.

\subsection{Validation and Test Objects}
\label{sec:validation_test_sets}
In machine learning, it is essential to have separate validation and test sets to ensure robust model development and evaluation. We use the validation set to tune hyperparameters and select the best-performing model. The test set is then used to evaluate how well the final model generalizes to unseen objects with different height profiles.

For both our validation and test objects (rigid and soft), we 3-D print objects with channels using a digital light processing (DLP) based 3-D printer (Pico2 HD@27, Asiga). We use channels specifically as they resemble wrinkle ridges and valleys on the skin. This printer was operated by a top-down DLP system with a digital mirror device (DMD) and a UV–LED light source operating at 385 nm. The rigid resin used in this work is GR-1 (Pro3-Dure). The soft resin used for 3-D printing comprises of epoxy aliphatic acrylate (EAA, Ebecryl 113, Allnex USA) and aliphatic urethane acrylate (AUD, Ebecryl 8413, Allnex, USA) in the weight ratio of 1:1. 2 wt. \% TPO (diphenyl(2,4,6-trimethylbenzoyl)phosphine oxide, Genocure TPO, RAHN USA Corp.) as photo-initiator and 0.15 wt. \% of Rhodamine dye add to the acrylate mixture in a hot water bath at 86 $^\circ$C with string at 300 rpm~\cite{patel_AM_SUV, Velez_JMEMS, Robosoft-Kim, Zefang_Dynamic_wrinkling}. Each layer was irradiated for 0.2 s, and layer thickness was 12 $\mu$m. The detailed printing parameters for rigid and soft are included in Table S1 in the Supporting Information. The printed structures were sonicated with isopropyl alcohol (IPA) for 3 min to remove the uncured resin and post cured in UV oven (UVP CL-1000 UV Oven) for 6 min.

For our validation set, we use 4 3-D-printed objects with straight channels as seen in Fig.~\ref{fig:validation_pipeline}. The channels are 500 micron wide with 500 micron between each channel. The 4 objects have designed channels of 24 micron, 48 micron, 72 micron, and 96 micron depths. While the 4 objects are designed to have channels of 24, 48, 72, and 96 micron depth, when printed these channels are not perfect due to error in the printing process. In order to get a ground truth estimate of the channel depth, we use the Keyence VK-X3100 3-D surface optical profilometer. The images were taken using the Nikon 10X Plan lens with a numerical aperture of 0.3, in the "Laser Confocal" scan mode using the "Standard" resolution, with a laser gamma correction value of 0.45 and 0\% offset. The profilometer has a resolution of 0.1nm with an accuracy of $\pm2\%$ of the measured value. We measure the height profile along 13 cross sections, evenly spaced across the surface of the piece. For each cross section, we apply a peak detection algorithm~\cite{du2006improved}. To improve the peak detection algorithm performance, we apply smoothing using a third order Savgol filter. We invert the cross section by multiplying all heights by -1 and then run our peak detection algorithm again to find the valleys. Lastly, we take the absolute value of all differences across all cross sections. We can then use all the differences between peaks and valleys across all 13 cross sections to find the ground truth mean and standard deviation of the channels for that object. 

For each of the 4 validation objects, we collect a GelSight reading in both the x and y directions as reconstruction performance may differ in the two axes due to uneven surface illumination; this is specifically exaggerated in the GelSight Mini sensor we utilized due to the lack of even angle and spacing between the three waveguides as seen in Fig.~\ref{fig:overview}A. For the 8 validation images collected, for each of the 4 test pieces in the 2 orientations (vertical and horizontal), we assess the height reconstruction performance similar to the process for calculating ground truth using the optical profilometer data. We take cross sections of the reconstructed height map at 100 pixel increments. We apply the same peak and valley detection process to find the mean and standard deviation in height between the peaks and valleys. 

For our test set, we use 4 3-D-printed objects with circular channels as seen in Fig.~\ref{fig:validation_pipeline}. The channels are 350 micron wide with 350 micron between each channel. The 4 test objects have designed channels of 12 micron, 36 micron, 60 micron, and 84 micron respectively. We follow a similar procedure to the validation test pieces for determining the ground truth channel height using the Keyence optical profilometer with the same settings, except that we use cross sections in a circular pattern at angle increments of 15 degrees. 

We collect a single GelSight reading with the test piece approximately centered in relation to the sensor. Next, we visually label the center of the test object in our GelSight image. We then follow a similar procedure to the validation test pieces, except that we use cross sections in a circular pattern at angle increments of 15 degrees. 

We collected all data points presented in this manuscript at a constant normal force of 19.62 Newtons (N). We additionally provide metrics for all validation and test pieces at various other forces: 4.91 N, 9.81 N, 14.71 N. These are included in Table S9 and S10 in the Supporting Information. We additionally provide metrics on 4 test pieces that are identical to the aforementioned 4 rigid circular test pieces but are printed with a soft resin, in order to show the deformation effects of pressing the gel on soft channel pieces, similar to skin. This data and analysis is provided in the Supporting Information Fig. S1 and Fig. S2.

\subsection{Wrinkle Detection and Height Estimation Algorithms}
\label{sec:wrinkle_estimation_methods}
We detect wrinkles in the 3-D height maps by identifying valleys (wrinkle troughs). To do this, we iterate through all $ (x, y) \in R $, a $1300 \times 1300$ grid of locations on the height map. For each location, we perform angled cuts passing through that location at angles of $-60^\circ$, $-30^\circ$, $0^\circ$ (vertical line), $30^\circ$, $60^\circ$, and $90^\circ$ (horizontal line). We extract the corresponding 1D height cross-sections from the height map.

For each of these cross-sections, to identify valleys, we invert the cross section by multiplying all height values by $-1$ and apply peak detection~\cite{du2006improved}. The resulting valleys correspond to wrinkle locations. Lastly, we skeletonize all valley candidate points using a 2-D skeletonization algorithm~\cite{zhang1984fast}.

Next, to estimate wrinkle depth, from our skeletonized valleys, we randomly sample $10{,}000$ valley points for time and compute efficiency. For each valley point, we search for the highest peak within a $30$ pixel radius, approximately $0.23$~mm. This value serves as a reasonable approximation informed by typical ridge-to-ridge distances observed in human fingerprints~\cite{moore1989analysis}. The height difference between each valley and its corresponding highest peak within this radius is computed as the wrinkle depth. Of the 10,000 total wrinkle height estimates, we use the 80th percentile value to represent wrinkle depth. This choice reflects the observation that many body locations contain numerous micro-wrinkles, which dominate the lowest points in the data, while deeper, more prominent wrinkles are less common. The 80th percentile provides a rough but more representative estimate of overall wrinkle depth. 

\subsection{Human Study Procedure}
\label{sec:human_study_procedure}
We conduct a human study with 15 participants, approved by Carnegie Mellon University's Institutional Review Board (Protocol Number: 2023\_00000269). The objective is to collect GelSight images to characterize wrinkle depths at various body locations and to examine the effects of an over-the-counter moisturizer on skin wrinkle depth.

We use a custom-designed 3-D-printed probe (Fig.~\ref{fig:overview}B) with an integrated load cell to collect GelSight readings with consistent force during skin contact. The load cell is a TE Connectivity FX1901-0001-0025-L, with a maximum operating force of approximately $111.21$ N and an accuracy of $\pm1\%$. Force signals from the load cell are amplified using the SparkFun Load Cell Amplifier Breakout Board with an Avia Semiconductor HX711 Analog to Digital Converter, and read by an Arduino Nano microcontroller. The Arduino transmits the force measurements to a companion laptop, which displays both live GelSight images and force data. A button embedded in the probe handle allows the researcher to tare (zero) the load cell, accounting for gravitational effects from probe orientation. The researcher tares before each indentation on the skin and monitors the live force and image streams to ensure proper contact. A GelSight image is captured by pressing a key on the keyboard of the companion laptop. All readings are taken at a load cell reading of approximately 19.62 N (2000 grams). 

The study proceeds as follows. After obtaining written informed consent, we begin by collecting demographic information. Three body locations (palm, wrist, elbow) are marked with a Sharpie marker. Participants rub the areas using a nail polish removing wipe containing acetone on each location in order to remove moisture from the skin for initial readings. This is due to the studies being conducted in June in a geographical location (Pittsburgh, Pennsylvania, USA) with a relative humidity that regularly exceeds 75\%. Next, two GelSight readings are captured at each site with the sensor aligned to the Sharpie mark. We apply 0.1 milliliters (mL) of Vanicream Moisturizing Cream for Sensitive Skin using a syringe to each location. Participants wear a glove to apply the cream to the three marked locations, rubbing it in until no visible residue remains. A timer is set for 8 minutes to allow the skin to absorb the moisturizer. Meanwhile, we collect a single GelSight reading at an additional 7 locations: forehead, inside elbow, inside upper arm, top of the middle finger, knuckle, and top of the hand. Once the 8 minutes have elapsed, the participant uses a disinfecting hand wipe to wipe any cream residue from the 3 moisturized areas. After the locations dry out, we realign the GelSight sensor with the Sharpie marks and capture 1 post-treatment image at each of the three locations.

All measurement locations are illustrated in Fig.\ref{fig:skin_various_locations}E and Fig.\ref{fig:before_after_cream}E. For all hand and wrist readings, participants rest their left hand in a relaxed position on a table. We sample the knuckle of the middle finger and the fingerprint region of the pinky finger. For the top of the finger location, we target the middle phalanx of the middle finger. The top of the hand is sampled in an area with minimal body hair. For the inside elbow and upper arm, participants fully extend their arm. For the outside elbow, participants bend their elbow and rest it on a table; the measurement is taken just above the joint, where the skin surface is flatter than directly over the joint.

\medskip
\textbf{Acknowledgments} \par 
This research was supported by the National Science Foundation Graduate Research Fellowship Program under Grant No. DGE1745016 and DGE2140739.

\medskip

%
\bibliographystyle{MSP}
\bibliography{bibliography}

\input{supporting_information.tex}

\end{document}

%% file: supporting_information.tex

\newpage
\section*{Supporting Information}
\subsubsection*{\phantomsection}
\section*{In-Vivo Skin 3-D Surface Reconstruction and Wrinkle Depth Estimation using Handheld High Resolution Tactile Sensing}
\setcounter{equation}{0}
\subsubsection*{\phantomsection}
\textit{Akhil Padmanabha, Arpit Agarwal, Catherine Li, Austin Williams, Dinesh K. Patel, Sankalp Chopkar, Achu Wilson, Ahmet Ozkan, Wenzhen Yuan, Sonal Choudhary, Arash Mostaghimi, Zackory Erickson, Carmel Majidi}

\newpage
\subsubsection*{\phantomsection}

\subsubsection*{Printing parameters}

\setcounter{table}{0}
\begin{table}[H]
    \centering
    \renewcommand{\arraystretch}{1.2}
    \renewcommand{\thetable}{S\arabic{table}}
    \caption{\textbf{Printing Parameters for Validation and Test Sets.}}
\label{table:S1}
\begin{center}
\begin{tabular}{||c|c|c||} 
 \hline
 \hline
 Printing parameter & Rigid & Soft \\ [0.5ex] 
 \hline\hline
 Slice thickness (mm) & 0.012 & 0.012 \\
 \hline
 Exposure time (s) & 0.3 & 0.3\\
 \hline
 Burn-in layers  & 3  & 3\\
 \hline
 Burn-in exposure time (s) & 6.0 & 6.0 \\
 \hline
 Light intensity (mW/cm$^2$) & 6.0 & 23.70\\
 \hline
 Heater temperature ($^\circ$C) & 30.0 & 40.0 \\
 \hline
 Separation velocity (mm/s) & 1.000 & 1.000\\
 \hline
 Separation distance (mm) & 6.000 & 6.000\\
 \hline
 Approach velocity (mm/s) & 2.000 & 2.000\\
 \hline
 Slides per layer & 1 & 1\\
 \hline
 Slide velocity (mm/s) & 10.000 & 10.000\\
 \hline
 Burn-in wait time (after exposure) & 5.000 & 5.000\\
 \hline
  Burn-in wait time (after separation) & 5.000 & 5.000\\
 \hline
  Burn-in wait time (after approach) & 3.000 & 3.000\\
 \hline
  Burn-in wait time (after slide) & 5.000 & 5.000\\
 \hline
  Normal wait time (after exposure) & 5.000 & 5.000\\
 \hline
 Normal wait time (after separation) & 5.000 & 5.000\\
 \hline
  Normal wait time (after approach) & 5.000 & 5.000\\
 \hline
  Normal wait time (after slide) & 5.000 & 5.000\\
 \hline
 \hline

\end{tabular}
\end{center}
   
\end{table}

\newpage
\subsubsection*{Device Comparison}

We provide a comparison of our method to commercially available devices in Table~\ref{tab:device_comparison}. For the commercially available devices, we calculated the values in the resolution column using the dimensions of the image and the dimensions of the sensitive area provided in the datasheets.

\begin{table}[H]
    \centering
    \renewcommand{\arraystretch}{1}
    \renewcommand{\thetable}{S\arabic{table}}
    
    \caption{\textbf{Comparison of our method with common commercially available skin topography devices.}}
\setlength{\tabcolsep}{4pt}
\begin{tabularx}{\textwidth}{|>{\raggedright\arraybackslash}X|>{\raggedright\arraybackslash}X|>{\raggedright\arraybackslash}X|c|c|>{\raggedright\arraybackslash}X|}
\hline
\textbf{Device} & \textbf{Method} & \textbf{Z-Depth Mean Absolute Error} & \textbf{Portable} & \textbf{Sensitive Area} & \textbf{Resolution (X $\times$ Y $\times$ Z)} \\ \hline

\textbf{Ours} (GelSight Mini from GelSight Inc. with Custom Gel) & Camera-Based Tactile Sensing (CNN + Poisson Integration) & $12.55 \pm 11.35$ µm & \checkmark & $\sim$10\,mm $\times$ 10\,mm & $\sim$7.7\,µm $\times$ $\sim$7.7\,µm $\times$ $<$1\,µm \\ \hline

Visioscan VC 20plus (Courage + Khazaka electronic GmbH)~\cite{courage2024visioscan} & UVA-light camera (Surface Evaluation of Living Skin) & Not Available & \checkmark & $\sim$10\,mm $\times$ 8\,mm & 7.8\,µm $\times$ 7.8\,µm $\times$ 15.6\text{--}50\,µm \\ \hline

Visioline VL 650 (Courage + Khazaka electronic GmbH)~\cite{courage2025visioline} & Silicone Replica and Desktop Angled Camera For Measuring Shadows & Not Available & $\times$ & 16.6\,mm $\times$ 22\,mm & 5.21\,µm $\times$ 9.375\,µm $\times$ 2\,µm \\ \hline

PRIMOS CR Small Field (Canfield)~\cite{primosCR_canfield} & Stereo and Fringe Optical Camera Based Reconstruction & 2\,µm (using a spherical test object) & \checkmark & 45\,mm $\times$ 30\,mm & 20\,µm $\times$ 20\,µm $\times$ 2\,µm \\ \hline

\end{tabularx}

\label{tab:device_comparison}
\end{table}

\newpage
\subsubsection*{Sensor Performance with Various Forces and with Soft Test Pieces}

\begin{figure}[h!]
  \centering
  \renewcommand{\thefigure}{S\arabic{figure}}
  \includegraphics[width = \textwidth]{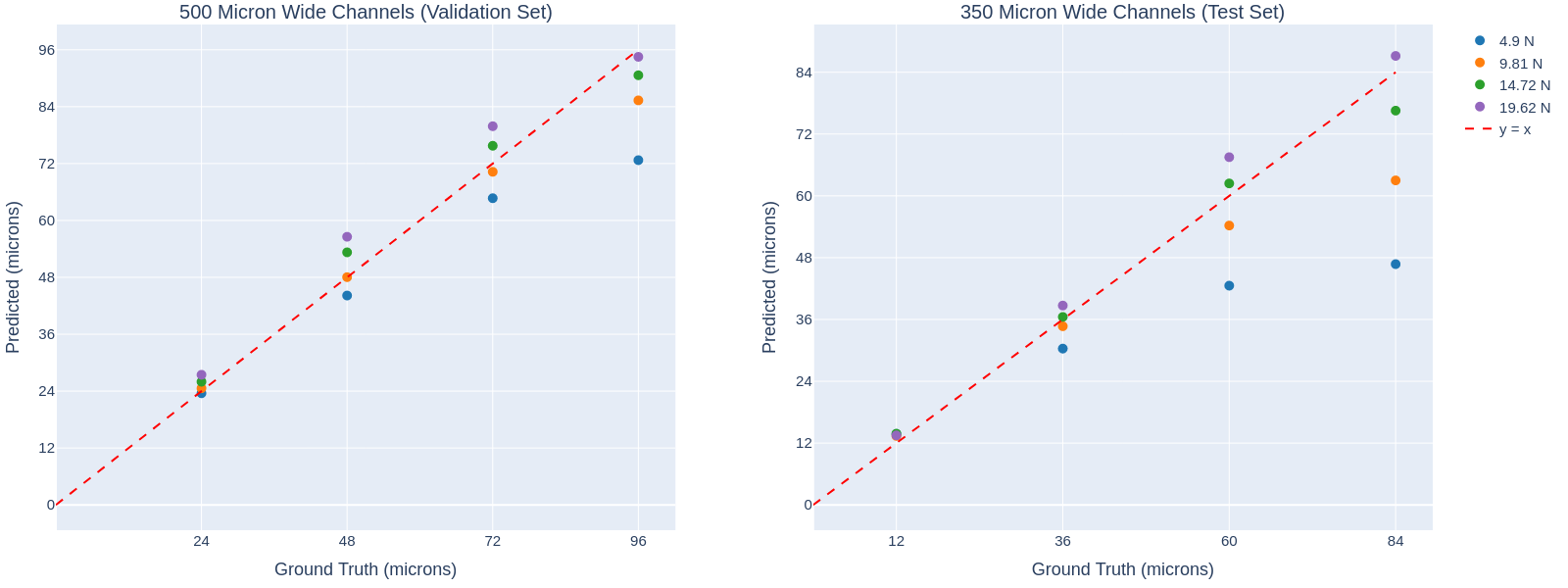}
  \caption{Left: Predicted wrinkle depth vs designed ground truth wrinkle depth for the validation set at 4 different forces. Right: Predicted wrinkle depth vs designed ground truth wrinkle depth for the test set at 4 different forces.}
  \label{fig:results_forces_plots}
\end{figure}

For the rigid validation and test sets, we collect readings at 4 forces: 4.91 N, 9.81 N, 14.71 N, and 19.62 N. We plot the predicted versus designed ground truth values at these forces in Fig.~\ref{fig:results_forces_plots}. We find that as less force is applied, the estimated depth reduces as a result of the lack of conformity of the gel to the channels. This is especially the case for the test set due to the more narrow channels of 350 micron width in comparison to the validation set which has 500 micron wide channels. The data shows that 19.62 N guarantees the conformity of the gel to both the validation and test set channels.

\begin{figure}[h!]
  \centering
  \renewcommand{\thefigure}{S\arabic{figure}}
  \includegraphics[width = \textwidth]{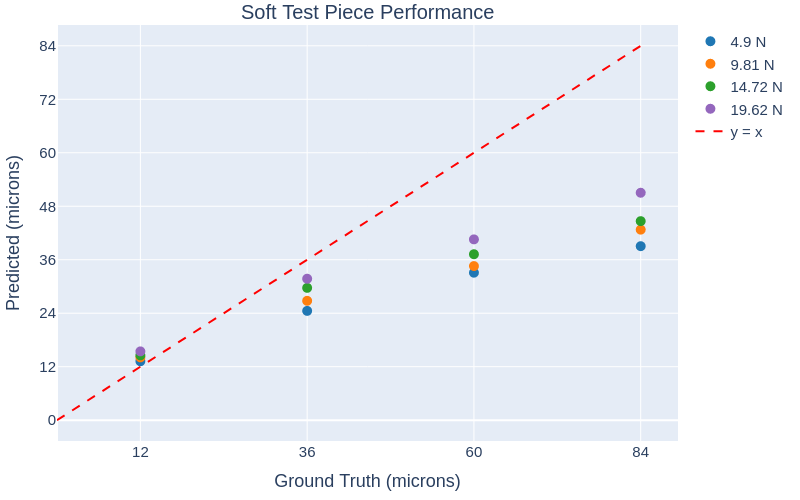}
  \caption{Predicted wrinkle depth vs designed ground truth wrinkle depth for the soft test set at 4 different forces.}
  \label{fig:soft_validation_performance}
\end{figure}

We also tested the sensor performance using a soft test piece. The soft peice was printed using a soft elastomeric resin with a Young's modulus of 0.6 MPa~\cite{patel_AM_SUV}. In Fig.~\ref{fig:soft_validation_performance}, we plot the predicted versus designed ground truth values at the 4 forces. We find that for smaller wrinkle depths of 12 and 36 micron, the gel is able to conform sufficiently. However, at larger wrinkle depths, we notice reduced predicted wrinkle depths due to deformation of the soft object. Future work should examine modeling this deformation and accounting for it in the predictions.

\newpage
\subsubsection*{Statistical Testing}
Results from statistical testing as described in the main text are shown in Table~\ref{tab:statistical_testing}.
\begin{table}[H]
    \centering
    \renewcommand{\arraystretch}{1.2}
    \renewcommand{\thetable}{S\arabic{table}}
    \caption{\textbf{Statistical testing results for skin locations before and after moisturizer application.}}
    \small
    \setlength{\tabcolsep}{3pt}
    \begin{adjustbox}{width=\textwidth}
    \begin{tabular}{|l|c|c|c|c|c|c|c|c|}
    \hline
    \makecell{\textbf{Location}} & 
    \makecell{\textbf{Friedman} \\ \textbf{Statistic}} & 
    \makecell{\textbf{Friedman} \\ \textbf{$p$-value}} & 
    \makecell{\textbf{Wilcoxon} \\ \textbf{Pre1 vs Pre2}} & 
    \makecell{\textbf{Wilcoxon} \\ \textbf{Pre1 vs Post}} & 
    \makecell{\textbf{Wilcoxon} \\ \textbf{Pre2 vs Post}} & 
    \makecell{\textbf{Wilcoxon with Bonf.} \\ \textbf{Pre1 vs Pre2}} & 
    \makecell{\textbf{Wilcoxon with Bonf.} \\ \textbf{Pre1 vs Post}} & 
    \makecell{\textbf{Wilcoxon with Bonf.} \\ \textbf{Pre2 vs Post}} \\
    \hline

    Palm & 19.7333 & 0.0001 & 0.5995 & 0.0001 & 0.0002 & 1.0000 & 0.0002 & 0.0005 \\
    Wrist & 22.5333 & 0.0000 & 0.2293 & 0.0001 & 0.0001 & 0.6879 & 0.0002 & 0.0002 \\
    Elbow & 21.7333 & 0.0000 & 0.1514 & 0.0001 & 0.0002 & 0.4543 & 0.0002 & 0.0005 \\
    \hline
    \end{tabular}
    \end{adjustbox}
    \label{tab:statistical_testing}
\end{table}

\newpage
\subsubsection*{Skin Wrinkle Depth Raw Data}
Raw data from our human studies is shown in Table~\ref{tab:skin_data_all_locations}, Table~\ref{tab:skin_data_cream}, and Table~\ref{tab:all_skin_wrinkle_raw_data_per_participant}.

\begin{table}[H]
    \centering
    \renewcommand{\thetable}{S\arabic{table}}
    \caption{\textbf{Skin Locations Boxplot Raw Data}}
    \small
    \vspace{-0.2cm}
\begin{tabular}{lrrrrr}
\toprule
Skin Location & Mean (\textmu m) & SD (\textmu m) & Median (\textmu m) & Min (\textmu m) & Max (\textmu m) \\
\midrule
Forehead & 14.88 & 4.10 & 13.58 & 10.47 & 22.21 \\
Upper Arm & 11.09 & 2.17 & 10.75 & 8.01 & 16.69 \\
Inside Elbow & 14.35 & 3.39 & 13.91 & 9.54 & 21.90 \\
Top Hand & 21.12 & 5.36 & 19.44 & 13.05 & 31.21 \\
Knuckle & 35.90 & 10.22 & 36.66 & 20.50 & 62.17 \\
Top Finger & 28.46 & 5.70 & 29.64 & 17.57 & 35.22 \\
Fingerprint & 21.54 & 4.42 & 22.88 & 12.50 & 28.28 \\
\bottomrule
\end{tabular}
\label{tab:skin_data_all_locations}
\end{table}

\begin{table}[H]
    \centering
    \renewcommand{\thetable}{S\arabic{table}}
    \caption{\textbf{Skin Locations Pre and Post Moisturizer Boxplot Raw Data}}
    \small
    \vspace{-0.2cm}
\begin{tabular}{lrrrrr}
\toprule
Skin Location & Mean (\textmu m) & SD (\textmu m) & Median (\textmu m) & Min (\textmu m) & Max (\textmu m) \\
\midrule
Palm Pre 1 & 19.10 & 4.98 & 18.26 & 10.12 & 27.41 \\
Palm Pre 2 & 18.77 & 4.80 & 17.59 & 10.07 & 27.14 \\
Palm Post & 13.90 & 3.59 & 13.38 & 8.80 & 21.83 \\
Wrist Pre 1 & 16.73 & 4.82 & 14.29 & 10.01 & 27.06 \\
Wrist Pre 2 & 16.34 & 4.93 & 14.06 & 10.22 & 27.08 \\
Wrist Post & 11.33 & 2.77 & 10.33 & 8.27 & 17.60 \\
Elbow Pre 1 & 29.88 & 6.87 & 28.54 & 20.58 & 47.46 \\
Elbow Pre 2 & 29.05 & 8.20 & 27.10 & 19.90 & 55.27 \\
Elbow Post & 18.57 & 5.71 & 16.87 & 10.94 & 28.40 \\
\bottomrule
\end{tabular}
\label{tab:skin_data_cream}
\end{table}

\begin{table}[H]
    \centering
    \renewcommand{\thetable}{S\arabic{table}}
    \caption{\textbf{80th Percentile Skin Wrinkle Depth per Participant per Location}}
    \label{tab:all_skin_wrinkle_raw_data_per_participant}
    \vspace{-0.2cm}
    \resizebox{\textwidth}{!}{%
    \begin{tabular}{lrrrrrrrrrrrrrrrr}
    \toprule
 & Forehead & Upper Arm & Inside Elbow & Top Hand & Knuckle & Top Finger & Fingerprint & Palm Pre 1 & Palm Pre 2 & Palm Post & Wrist Pre 1 & Wrist Pre 2 & Wrist Post & Elbow Pre 1 & Elbow Pre 2 & Elbow Post \\
Participant ID &  &  &  &  &  &  &  &  &  &  &  &  &  &  &  &  \\
\midrule
0 & 21.75 & 12.08 & 15.08 & 24.30 & 62.17 & 34.82 & 18.41 & 16.07 & 15.48 & 8.80 & 19.04 & 20.70 & 13.73 & 20.58 & 20.56 & 15.38 \\
1 & 13.58 & 16.69 & 21.90 & 31.21 & 43.99 & 35.19 & 24.27 & 26.96 & 25.73 & 17.47 & 21.85 & 21.05 & 11.87 & 28.23 & 26.11 & 12.00 \\
2 & 15.71 & 9.62 & 10.37 & 28.01 & 37.28 & 26.65 & 24.35 & 13.44 & 14.47 & 11.17 & 12.41 & 11.91 & 9.21 & 28.70 & 24.93 & 27.28 \\
3 & 11.57 & 10.75 & 16.20 & 24.83 & 42.32 & 35.22 & 25.28 & 18.26 & 15.76 & 17.25 & 13.23 & 14.06 & 10.50 & 28.10 & 30.10 & 26.75 \\
4 & 20.13 & 13.10 & 16.90 & 22.90 & 36.20 & 34.50 & 26.00 & 19.95 & 21.29 & 11.98 & 19.38 & 19.87 & 9.58 & 47.46 & 55.27 & 12.77 \\
5 & 12.24 & 9.95 & 13.91 & 30.03 & 30.30 & 25.86 & 21.56 & 10.12 & 10.07 & 10.02 & 12.92 & 13.20 & 8.27 & 30.59 & 30.57 & 28.40 \\
6 & 20.76 & 11.51 & 18.48 & 18.49 & 42.50 & 30.48 & 24.19 & 24.06 & 24.32 & 14.11 & 27.06 & 27.08 & 13.30 & 24.24 & 24.32 & 19.09 \\
7 & 14.45 & 10.68 & 14.90 & 17.72 & 41.20 & 29.64 & 16.43 & 18.72 & 18.92 & 10.41 & 16.97 & 14.83 & 9.72 & 30.05 & 27.10 & 10.94 \\
8 & 10.76 & 8.08 & 9.55 & 15.24 & 20.50 & 17.57 & 20.92 & 14.21 & 13.99 & 11.65 & 13.25 & 10.73 & 8.90 & 27.57 & 30.35 & 20.41 \\
9 & 13.34 & 13.86 & 17.70 & 20.80 & 27.61 & 32.06 & 24.89 & 27.41 & 24.05 & 18.74 & 23.33 & 21.39 & 17.60 & 33.81 & 31.56 & 22.79 \\
10 & 22.21 & 11.58 & 13.20 & 19.44 & 36.66 & 26.98 & 28.28 & 26.08 & 27.14 & 21.83 & 21.16 & 21.30 & 15.81 & 42.56 & 37.23 & 23.28 \\
11 & 10.47 & 9.40 & 13.44 & 13.05 & 25.84 & 18.30 & 22.88 & 15.68 & 14.54 & 10.93 & 11.72 & 11.19 & 10.33 & 21.21 & 19.90 & 14.85 \\
12 & 10.88 & 8.01 & 9.54 & 17.76 & 21.87 & 20.55 & 12.50 & 16.69 & 17.59 & 15.55 & 10.01 & 10.22 & 8.84 & 28.54 & 24.64 & 13.19 \\
13 & 11.37 & 10.02 & 11.49 & 16.91 & 29.55 & 30.39 & 14.24 & 20.71 & 20.78 & 15.24 & 14.29 & 13.77 & 13.63 & 26.25 & 25.16 & 16.87 \\
14 & 13.93 & 11.03 & 12.53 & 16.06 & 40.45 & 28.68 & 18.92 & 18.16 & 17.47 & 13.38 & 14.28 & 13.73 & 8.62 & 30.36 & 27.90 & 14.60 \\
Means & 14.88 & 11.09 & 14.35 & 21.12 & 35.90 & 28.46 & 21.54 & 19.10 & 18.77 & 13.90 & 16.73 & 16.34 & 11.33 & 29.88 & 29.05 & 18.57 \\
\bottomrule
\end{tabular}
}
\end{table}

\newpage
\subsubsection*{Validation and Test Object Data for Gel 1 and Gel 2}
All raw data used for calculating metrics and for plots in the main text are provided for Gel 1 and Gel 2 in Table~\ref{tab:validation_object_average_data}, Table~\ref{tab:test_object_average_data}, Table~\ref{tab:Gel1_data}, and Table~\ref{tab:Gel2_data}.

\begin{table}[h!]
    \centering
    \renewcommand{\thetable}{S\arabic{table}}
    \caption{\textbf{Mean and SD for every Validation Object Averaged Across Gel 1 and Gel 2}}
    \small
    \vspace{-0.2cm}
    \begin{adjustbox}{width=\textwidth}
    \begin{tabular}{ccccc}
    \toprule
    Designed Ground Truth (\textmu m) & Profilometer Mean (\textmu m) & Profilometer SD (\textmu m) & GelSight Mean (\textmu m) & GelSight SD (\textmu m) \\
    \midrule
    24 & 22.65 & 3.30 & 27.45 & 10.36 \\
    48 & 53.95 & 2.80 & 56.56 & 14.51 \\
    72 & 70.31 & 4.45 & 79.88 & 16.92 \\
    96 & 101.22 & 9.19 & 94.51 & 21.13 \\
    \bottomrule
    \end{tabular}
    \end{adjustbox}
    \label{tab:validation_object_average_data}
\end{table}

\begin{table}[h!]
    \centering
    \renewcommand{\thetable}{S\arabic{table}}
    \caption{\textbf{Mean and SD for every Test Object Averaged Across Gel 1 and Gel 2}}
    \small
    \vspace{-0.2cm}
    \begin{adjustbox}{width=\textwidth}
    \begin{tabular}{ccccc}
    \toprule
    Designed Ground Truth (\textmu m) & Profilometer Mean (\textmu m) & Profilometer SD (\textmu m) & GelSight Mean (\textmu m) & GelSight SD (\textmu m) \\
    \midrule
    12 & 9.09  & 1.35 & 13.50 & 7.96  \\
    36 & 35.05 & 2.86 & 38.70 & 12.46 \\
    60 & 66.92 & 2.95 & 67.51 & 13.96 \\
    84 & 92.34 & 6.90 & 87.17 & 24.10 \\
    \bottomrule
    \end{tabular}
    \end{adjustbox}
    \label{tab:test_object_average_data}
\end{table}

\newpage

\begin{table}[h!]
    \centering
    \renewcommand{\thetable}{S\arabic{table}}
    \renewcommand{\arraystretch}{0.65}
    \caption{\textbf{Gel 1 Validation and Test Piece Data}}
    \small
    
    \vspace{-0.3cm}
    \begin{adjustbox}{width=\textwidth}
    \begin{tabular}{ccccccc}
    \toprule
Dataset & Configuration & Firmness & Force (N) & Designed Ground Truth (\textmu m) & Mean (\textmu m) & SD (\textmu m) \\
\midrule
Validation & horizontal & hard & 4.91 & 96 & 66.33 & 24.08 \\
Validation & horizontal & hard & 9.81 & 96 & 84.46 & 21.89 \\
Validation & horizontal & hard & 14.71 & 96 & 94.47 & 23.98 \\
Validation & horizontal & hard & 19.62 & 96 & 103.02 & 25.98 \\
Validation & vertical & hard & 4.91 & 96 & 75.01 & 20.91 \\
Validation & vertical & hard & 9.81 & 96 & 90.99 & 17.52 \\
Validation & vertical & hard & 14.71 & 96 & 94.25 & 15.68 \\
Validation & vertical & hard & 19.62 & 96 & 95.22 & 15.12 \\
Validation & horizontal & hard & 4.91 & 72 & 58.63 & 14.16 \\
Validation & horizontal & hard & 9.81 & 72 & 67.66 & 14.25 \\
Validation & horizontal & hard & 14.71 & 72 & 76.12 & 15.18 \\
Validation & horizontal & hard & 19.62 & 72 & 85.11 & 17.91 \\
Validation & vertical & hard & 4.91 & 72 & 74.32 & 15.07 \\
Validation & vertical & hard & 9.81 & 72 & 76.93 & 14.98 \\
Validation & vertical & hard & 14.71 & 72 & 82.50 & 13.73 \\
Validation & vertical & hard & 19.62 & 72 & 85.55 & 14.56 \\
Validation & horizontal & hard & 4.91 & 48 & 36.91 & 11.96 \\
Validation & horizontal & hard & 9.81 & 48 & 44.14 & 13.05 \\
Validation & horizontal & hard & 14.71 & 48 & 52.36 & 13.88 \\
Validation & horizontal & hard & 19.62 & 48 & 58.13 & 12.41 \\
Validation & vertical & hard & 4.91 & 48 & 47.57 & 13.87 \\
Validation & vertical & hard & 9.81 & 48 & 48.59 & 13.62 \\
Validation & vertical & hard & 14.71 & 48 & 52.87 & 12.59 \\
Validation & vertical & hard & 19.62 & 48 & 54.68 & 13.11 \\
Validation & horizontal & hard & 4.91 & 24 & 17.70 & 7.24 \\
Validation & horizontal & hard & 9.81 & 24 & 19.43 & 7.78 \\
Validation & horizontal & hard & 14.71 & 24 & 22.53 & 8.03 \\
Validation & horizontal & hard & 19.62 & 24 & 26.73 & 8.93 \\
Validation & vertical & hard & 4.91 & 24 & 25.18 & 11.44 \\
Validation & vertical & hard & 9.81 & 24 & 26.12 & 10.36 \\
Validation & vertical & hard & 14.71 & 24 & 24.88 & 10.38 \\
Validation & vertical & hard & 19.62 & 24 & 24.97 & 9.62 \\
Test & circle & hard & 4.91 & 84 & 47.99 & 22.02 \\
Test & circle & hard & 9.81 & 84 & 67.87 & 24.40 \\
Test & circle & hard & 14.71 & 84 & 83.57 & 28.10 \\
Test & circle & hard & 19.62 & 84 & 93.22 & 29.02 \\
Test & circle & hard & 4.91 & 60 & 43.80 & 18.51 \\
Test & circle & hard & 9.81 & 60 & 56.50 & 15.74 \\
Test & circle & hard & 14.71 & 60 & 66.83 & 13.74 \\
Test & circle & hard & 19.62 & 60 & 73.87 & 13.77 \\
Test & circle & hard & 4.91 & 36 & 28.29 & 13.57 \\
Test & circle & hard & 9.81 & 36 & 35.04 & 13.73 \\
Test & circle & hard & 14.71 & 36 & 37.55 & 13.07 \\
Test & circle & hard & 19.62 & 36 & 40.68 & 12.99 \\
Test & circle & hard & 4.91 & 12 & 13.71 & 10.17 \\
Test & circle & hard & 9.81 & 12 & 13.60 & 9.95 \\
Test & circle & hard & 14.71 & 12 & 13.51 & 9.53 \\
Test & circle & hard & 19.62 & 12 & 13.95 & 9.15 \\
Test & circle & soft & 4.91 & 84 & 36.06 & 10.42 \\
Test & circle & soft & 9.81 & 84 & 39.03 & 12.40 \\
Test & circle & soft & 14.71 & 84 & 40.47 & 14.31 \\
Test & circle & soft & 19.62 & 84 & 47.63 & 14.27 \\
Test & circle & soft & 4.91 & 60 & 33.74 & 12.59 \\
Test & circle & soft & 9.81 & 60 & 33.15 & 13.38 \\
Test & circle & soft & 14.71 & 60 & 35.69 & 13.91 \\
Test & circle & soft & 19.62 & 60 & 38.28 & 14.93 \\
Test & circle & soft & 4.91 & 36 & 21.75 & 10.73 \\
Test & circle & soft & 9.81 & 36 & 23.91 & 10.11 \\
Test & circle & soft & 14.71 & 36 & 27.02 & 9.97 \\
Test & circle & soft & 19.62 & 36 & 29.80 & 10.80 \\
Test & circle & soft & 4.91 & 12 & 11.52 & 8.89 \\
Test & circle & soft & 9.81 & 12 & 12.42 & 8.50 \\
Test & circle & soft & 14.71 & 12 & 12.59 & 8.54 \\
Test & circle & soft & 19.62 & 12 & 13.49 & 8.82 \\
  \bottomrule
    \end{tabular}
    \end{adjustbox}
    \label{tab:Gel1_data}
\end{table}

\begin{table}[h!]
    \centering
    \renewcommand{\thetable}{S\arabic{table}}
    \renewcommand{\arraystretch}{0.65}
    \caption{\textbf{Gel 2 Validation and Test Piece Data}}
    \small
    
    \vspace{-0.3cm}
    \begin{adjustbox}{width=\textwidth}
    \begin{tabular}{ccccccc}
    \toprule
Dataset & Configuration & Firmness & Force (N) & Designed Ground Truth (\textmu m) & Mean (\textmu m) & SD (\textmu m) \\
\midrule
Validation & horizontal & hard & 4.91 & 96 & 71.64 & 22.38 \\
Validation & horizontal & hard & 9.81 & 96 & 82.56 & 17.72 \\
Validation & horizontal & hard & 14.71 & 96 & 86.22 & 18.61 \\
Validation & horizontal & hard & 19.62 & 96 & 89.58 & 18.58 \\
Validation & vertical & hard & 4.91 & 96 & 77.88 & 24.05 \\
Validation & vertical & hard & 9.81 & 96 & 83.31 & 23.99 \\
Validation & vertical & hard & 14.71 & 96 & 87.56 & 24.75 \\
Validation & vertical & hard & 19.62 & 96 & 90.23 & 24.83 \\
Validation & horizontal & hard & 4.91 & 72 & 57.89 & 14.44 \\
Validation & horizontal & hard & 9.81 & 72 & 66.10 & 13.64 \\
Validation & horizontal & hard & 14.71 & 72 & 72.80 & 14.71 \\
Validation & horizontal & hard & 19.62 & 72 & 76.78 & 16.81 \\
Validation & vertical & hard & 4.91 & 72 & 67.93 & 19.23 \\
Validation & vertical & hard & 9.81 & 72 & 70.35 & 20.01 \\
Validation & vertical & hard & 14.71 & 72 & 71.60 & 19.43 \\
Validation & vertical & hard & 19.62 & 72 & 72.07 & 18.38 \\
Validation & horizontal & hard & 4.91 & 48 & 40.28 & 14.58 \\
Validation & horizontal & hard & 9.81 & 48 & 46.34 & 14.25 \\
Validation & horizontal & hard & 14.71 & 48 & 51.11 & 14.92 \\
Validation & horizontal & hard & 19.62 & 48 & 55.58 & 17.23 \\
Validation & vertical & hard & 4.91 & 48 & 51.89 & 14.15 \\
Validation & vertical & hard & 9.81 & 48 & 53.09 & 14.55 \\
Validation & vertical & hard & 14.71 & 48 & 56.76 & 15.57 \\
Validation & vertical & hard & 19.62 & 48 & 57.84 & 15.30 \\
Validation & horizontal & hard & 4.91 & 24 & 22.32 & 11.12 \\
Validation & horizontal & hard & 9.81 & 24 & 22.94 & 12.44 \\
Validation & horizontal & hard & 14.71 & 24 & 25.21 & 12.77 \\
Validation & horizontal & hard & 19.62 & 24 & 26.75 & 13.06 \\
Validation & vertical & hard & 4.91 & 24 & 29.00 & 9.47 \\
Validation & vertical & hard & 9.81 & 24 & 29.93 & 9.61 \\
Validation & vertical & hard & 14.71 & 24 & 31.28 & 9.37 \\
Validation & vertical & hard & 19.62 & 24 & 31.34 & 9.83 \\
Test & circle & hard & 4.91 & 84 & 45.50 & 12.01 \\
Test & circle & hard & 9.81 & 84 & 58.13 & 19.64 \\
Test & circle & hard & 14.71 & 84 & 69.52 & 20.24 \\
Test & circle & hard & 19.62 & 84 & 81.12 & 19.17 \\
Test & circle & hard & 4.91 & 60 & 41.33 & 14.69 \\
Test & circle & hard & 9.81 & 60 & 51.97 & 14.28 \\
Test & circle & hard & 14.71 & 60 & 57.99 & 13.64 \\
Test & circle & hard & 19.62 & 60 & 61.15 & 14.14 \\
Test & circle & hard & 4.91 & 36 & 32.40 & 11.73 \\
Test & circle & hard & 9.81 & 36 & 34.33 & 11.73 \\
Test & circle & hard & 14.71 & 36 & 35.44 & 11.57 \\
Test & circle & hard & 19.62 & 36 & 36.73 & 11.92 \\
Test & circle & hard & 4.91 & 12 & 13.90 & 6.77 \\
Test & circle & hard & 9.81 & 12 & 13.18 & 6.49 \\
Test & circle & hard & 14.71 & 12 & 14.12 & 6.78 \\
Test & circle & hard & 19.62 & 12 & 13.06 & 6.76 \\
Test & circle & soft & 4.91 & 84 & 42.03 & 14.91 \\
Test & circle & soft & 9.81 & 84 & 46.47 & 15.39 \\
Test & circle & soft & 14.71 & 84 & 48.85 & 15.25 \\
Test & circle & soft & 19.62 & 84 & 54.43 & 14.06 \\
Test & circle & soft & 4.91 & 60 & 32.45 & 14.74 \\
Test & circle & soft & 9.81 & 60 & 36.01 & 15.16 \\
Test & circle & soft & 14.71 & 60 & 38.79 & 15.09 \\
Test & circle & soft & 19.62 & 60 & 42.86 & 14.55 \\
Test & circle & soft & 4.91 & 36 & 27.30 & 11.32 \\
Test & circle & soft & 9.81 & 36 & 29.66 & 11.99 \\
Test & circle & soft & 14.71 & 36 & 32.36 & 11.72 \\
Test & circle & soft & 19.62 & 36 & 33.68 & 12.05 \\
Test & circle & soft & 4.91 & 12 & 14.94 & 7.06 \\
Test & circle & soft & 9.81 & 12 & 15.77 & 7.29 \\
Test & circle & soft & 14.71 & 12 & 16.50 & 7.60 \\
Test & circle & soft & 19.62 & 12 & 17.39 & 8.18 \\
  \bottomrule
    \end{tabular}
    \end{adjustbox}
    \label{tab:Gel2_data}
\end{table}